\documentclass[10pt,final,twocolumn,letterpaper,pagenumbers]{article}
\usepackage{cvpr}              

\usepackage{url}
\usepackage{graphicx}
\usepackage{tabularx}
\usepackage{comment}
\usepackage{makecell}
\usepackage{pifont}
\usepackage{multirow}
\usepackage[misc]{ifsym}
\usepackage{afterpage}
\usepackage{setspace}
\usepackage{breakcites}
\usepackage{seqsplit}
\usepackage{booktabs}
\usepackage{times}
\usepackage{epsfig}
\usepackage{amsmath}
\usepackage{amssymb}
\usepackage{float}
\usepackage[symbol]{footmisc}
\usepackage[title]{appendix}
\usepackage{lipsum}
\usepackage[table]{xcolor}
\usepackage{hyphenat}
\usepackage{soul,color}
\usepackage{amsopn}

\definecolor{Red}{cmyk}{0,1,1,0}
\definecolor{Green}{cmyk}{1,0,1,0}
\definecolor{Cyan}{cmyk}{1,0,0,0}
\definecolor{Purple}{cmyk}{0.45,0.86,0,0}
\definecolor{Rosolic}{cmyk}{0.00,1.00,0.50,0}
\definecolor{Blue}{cmyk}{1.00,1.00,0.00,0}
\definecolor{BlueViolet}{cmyk}{0.86,0.91,0,0.04}
\definecolor{NavyBlue}{cmyk}{0.94,0.54,0,0}

\newcommand{\hidden}[1]{{\color{NavyBlue}}}

\newcommand{\myparagraph}[1]{\vspace{0.3em}\noindent\textbf{#1}}
\newcommand{\loc}[1]{\textcolor[RGB]{228,26,28}{$<$loc$>$#1$<$/loc$>$}}
\newcommand{\obj}[1]{\textcolor[RGB]{55,126,184}{$<$obj$>$#1$<$/obj$>$}}
\newcommand{\sep}[1]{\textcolor[RGB]{255,127,0}{#1}}
\newcommand{\better}[1]{\textcolor[RGB]{77,175,74}{#1}}

\newcommand{\modelname}[1]{LL3DA}

\definecolor{cvprblue}{rgb}{0.21,0.49,0.74}
\usepackage[pagebackref,breaklinks,colorlinks,citecolor=cvprblue]{hyperref}
\usepackage[capitalize]{cleveref}
\crefname{section}{Sec.}{Secs.}
\Crefname{section}{Section}{Sections}
\Crefname{table}{Table}{Tables}
\crefname{table}{Tab.}{Tabs.}

\begin{document}

\title{
    LL3DA: Visual Interactive Instruction Tuning for \\
    Omni-3D Understanding, Reasoning, and Planning
}

\author{
    Sijin Chen$^{1}$ \quad
    Xin Chen$^{2,*}$ \quad
    Chi Zhang$^{2}$ \quad
    Mingsheng Li$^{1}$ \quad
    Gang YU$^{2}$ \quad
    \\
    Hao Fei$^{3}$ \quad
    Hongyuan Zhu$^{4}$ \quad
    Jiayuan Fan$^{1}$ \quad
    Tao Chen$^{1,\dagger}$
    \\
    {\normalsize $^{1}$Fudan University} \quad
    {\normalsize $^{2}$Tencent PCG} \quad
    {\normalsize $^{3}$National University of Singapore} \\
    {\normalsize $^{4}$Institute for Infocomm Research (I$^2$R) \&
     Centre for Frontier AI Research (CFAR), A*STAR, Singapore}
    \\
    \tt \small \textbf{\href{https://github.com/Open3DA/LL3DA}{https://github.com/Open3DA/LL3DA}}
    \\
    {\small $^{*}$ project lead} \quad {\small $^{\dagger}$ corresponding author}
}


\twocolumn[{
    \renewcommand\twocolumn[1][]{#1}
    \maketitle
    \begin{center}
        \captionsetup{type=figure}
        \includegraphics[width=\textwidth]{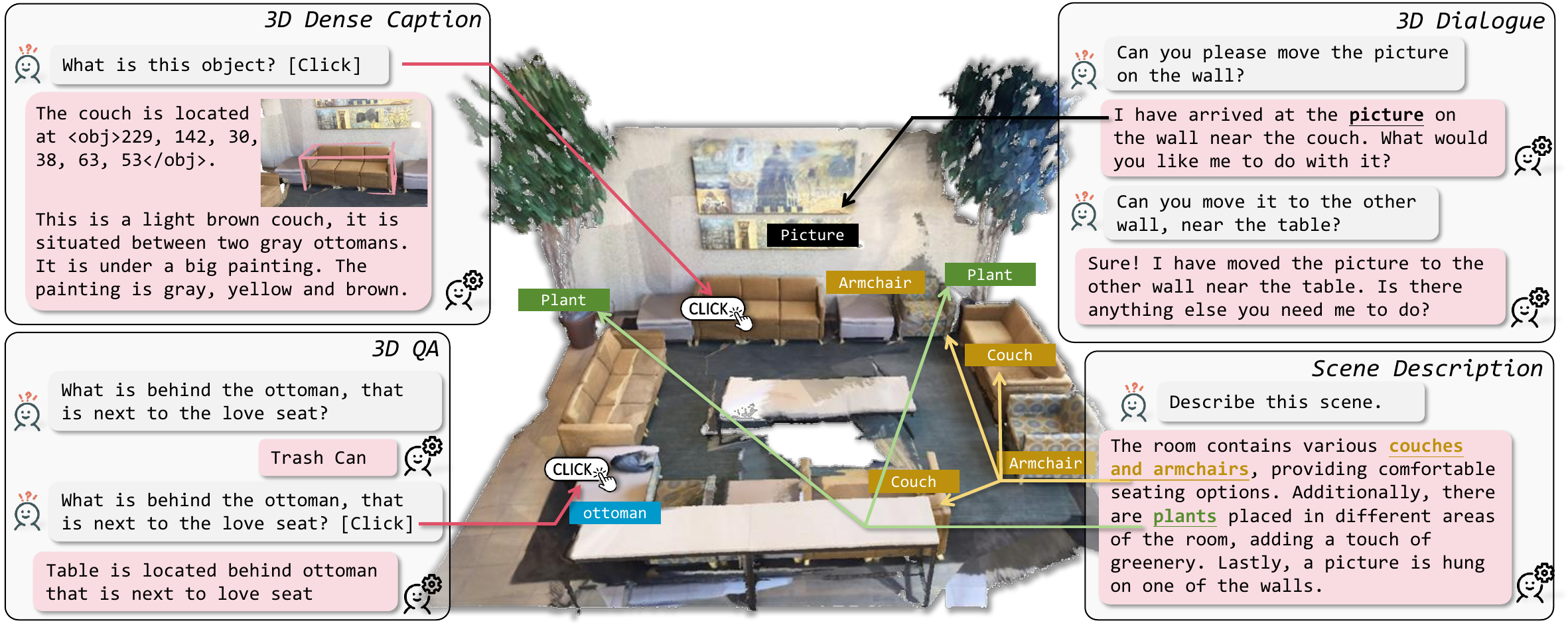}
        \caption{
            \textbf{We propose \modelname{}, a Large Language 3D Assistant that demonstrates mighty instruction-following capacities of understanding, reasoning, and planning in complex 3D environments.}
            \modelname{} takes both the textual instructions and potential visual interactions into consideration to help remove ambiguities when addressing various tasks in diverse 3D scenes.
        }
        \label{fig:teaser}
    \end{center}
}]

\begin{abstract}

Recent advances in \textbf{L}arge \textbf{M}ultimodal \textbf{M}odels (LMM) have made it possible for various applications in human-machine interactions.
However, developing LMMs that can comprehend, reason, and plan in complex and diverse 3D environments remains a challenging topic, especially considering the demand for understanding permutation-invariant point cloud 3D representations of the 3D scene.
Existing works seek help from multi-view images, and project 2D features to 3D space as 3D scene representations.
This, however, leads to huge computational overhead and performance degradation.
In this paper, we present \modelname{}, a \textbf{L}arge \textbf{L}anguage \textbf{3D} \textbf{A}ssistant that takes point cloud as direct input and respond to both textual-instructions and visual-prompts.
This help LMMs better comprehend human interactions and further help to remove the ambiguities in cluttered 3D scenes.
Experiments show that \modelname{} achieves remarkable results, and surpasses various 3D vision-language models on both 3D Dense Captioning and 3D Question Answering.
%

\end{abstract}

\vspace{-4mm}
\section{Introduction}
\label{sec:intro}

\hspace{1em}
The recent surge in \textbf{L}arge \textbf{L}anguage \textbf{M}odel (LLM) families \cite{touvron2023llama,zhang2022opt,chung2022flan-t5,openai2023gpt4,iyer2022opt-iml} opens up great opportunities for solving various machine learning tasks in a generalized way \cite{liu2023llava,li2023blip2,instructblip,jiang2023motiongpt}.
During this LLM carnival, researchers are also seeking generalized LLM solutions to various vision language tasks \cite{instructblip,xu2023pointllm,zhang2023gpt4roi}.
Among these, LLM-based 3D scene understanding is a valuable topic that would benefit the development of autonomous driving \cite{fu2023drive,chen2023driving} and embodied AI agents \cite{driess2023palm,song2023llm}.
However, it is also challenging given 1) the diversity and complication of 3D environments and 2) the demands for understanding sparse 3D points.

Prior works have made initial success addressing various 3D vision and language tasks.
The mainstream of researches build 3D specialists aiming at solving one specific down-stream task, including 
\textbf{3D} \textbf{Q}uestion \textbf{A}nswering (3D-QA) \cite{azuma2022scanqa,ma2022sqa3d}, 
\textbf{3D} \textbf{V}isual \textbf{G}rounding (3D-VG) \cite{chen2020scanrefer,wu2023eda}, and 
\textbf{3D} \textbf{D}ense \textbf{C}aptioning (3D-DC) \cite{chen2021scan2cap,chen2023vote2cap-detr}.
Meanwhile, other works \cite{zhu20233d-vista,jin2023-3D-VLP,cai20223djcg,chen2023unit3d} study the mutual promotion of different 3D vision and language tasks with shared structure modelling relations among objects.
Recently, researchers have also introduced LLMs for general purpose 3D understanding, where Point-Bind and Point-LLMs \cite{xu2023pointllm,guo2023pointbind+pointllm} mainly focus on the understanding of 3D objects.
Concurrently, 3D-LLM \cite{hong20233d-llm} proposes an LLM-driven solution that aggregates multi-view features for 3D features, presenting mighty capacities a machine could understand various 3D object and scenes, and follow textual instructions produced by human.

Though these methods have achieved remarkable success addressing different challenges in understanding 3D worlds with natural language, there are certain limitations.
With limited supervision, 3D specialists could hardly scale-up for better performance, while the joint pre-training still requires separate heads for specific tasks.
Extracting mutli-view features results in huge computational overhead, and ignores the essential geometry properties.
Additionally, plain texts often lead to ambiguities especially in cluttered and complex 3D environments.



%
%

To address the above issues, we propose the \modelname{}, a \textbf{L}arge \textbf{L}anguage \textbf{3D} \textbf{A}ssistant that could respond to both textual and visual interactions from human, and understand, reason, and plan in complex 3D environments (\cref{fig:teaser}).
%
%
We adopt a multi-modal transformer that aggregates information from textual instructions, visual prompts, and 3D scene into a fixed length of learnable querying tokens via the attention mechanism.
The querying tokens are projected and used as the prefix of the textual instructions, serving as the input to a pre-trained and frozen LLM.
This design not only helps to address the contradiction between the permutation-invariant 3D scene embeddings with the LLM embedding space, but also extracts interaction-aware 3D scene embeddings for efficient instruction following.

We conduct extensive experiments to explore the capacities of LLMs in understanding, reasoning, and planning within complex and diverse 3D environments.
Our model achieves state-of-the-art results on two widely used datasets for 3D Dense Captioning \cite{chen2020scanrefer,achlioptas2020referit3d}, and 3D Question Answering \cite{azuma2022scanqa}.
Additionally, by introducing additional visual interactions, our method could further remove the ambiguities within the vague textual instructions.

To summarize, our key contributions lie in:

\begin{itemize} 
\setlength\itemsep{0em}

    \item We present a LLM-based solution for understanding, reasoning, and planning in complex 3D environments.
    
    \item Our model takes both the textual instructions and visual interactions as inputs, and extracts interaction-aware features for effective instruction-following.

    \item Extensive experiments show that our method surpasses various state-of-the-art 3D vision language models.
    
\end{itemize}
\section{Related Work}
\label{sec:related}

\myparagraph{3D Vision and Language} alignment, pre-training, and understanding \cite{ding2023pla,zhu20233d-vista,chen2020scanrefer} covers a bunch of tasks requiring a model to adopt its understanding towards a complex 3D scene answering to, or answering with natural language.
Among those, \textbf{3D} \textbf{D}ense \textbf{C}aptioning (3D-DC) \cite{chen2021scan2cap,wang2022spacap3d,chen2023vote2cap-detr} expects a model to translate an input 3D scene into a set of instance coordinates and natural language descriptions.
Existing methods could be categorized into ``detect-then-describe'' models \cite{cai20223djcg,chen2021scan2cap,wang2022spacap3d} and the ``set-to-set'' prediction approaches \cite{chen2023vote2cap-detr,chen2023vote2cap-detr++}.
The former builds explicit relations on the instance coordinate predictions, while the latter direct learns the location and description of instances from the input 3D scene.
\textbf{3D} \textbf{V}isual \textbf{G}rounding (3D-VG) \cite{chen2020scanrefer,achlioptas2020referit3d,wu2023eda} demands a model to respond the natural language queries with the instance coordinates in the 3D scene.
The mainstream of existing methods \cite{cai20223djcg,zhu20233d-vista,zhao20213dvg} address 3D-VG via selecting a candidate from a 3D detector's prediction.
\textbf{3D} \textbf{Q}uestion \textbf{A}nswering (3D-QA) \cite{ye20223d,azuma2022scanqa,ma2022sqa3d,zhao2022-fe-3dgqa} requires a model to answer the questions with natural language based on the input 3D scene.
The majority of existing methods \cite{azuma2022scanqa,delitzas2023multi-clip,parelli2023clip-guided} directly select the desired response from a given answer set.
Researchers have also studied the mutual promotion of various 3D vision language tasks via training their shareable architectures simultaneously on different tasks \cite{cai20223djcg,chen2023unit3d,zhu20233d-vista,jin2023-3D-VLP}.
UniT3D \cite{chen2023unit3d} and 3DJCG \cite{cai20223djcg} focus on the joint promotion between 3D-DC and 3D-VG in the relation modelling, while 3D-VLP \cite{jin2023-3D-VLP} further includes 3D-QA.
Recently, 3D-LLM \cite{hong20233d-llm} introduce a family of LLM-driven 3D generalists that could handle diverse textual instructions with the reconstructed 3D feature from multi-view images \cite{hong20233d}.
In this paper, we present \modelname{}, an LLM solution that directly extracts features from the 3D scene, and could handle both visual prompts and textual instructions to diversify the possible interactions human could make with the complex 3D environment.

\begin{figure*}[htbp]
	\centering
	\includegraphics[width=\linewidth]{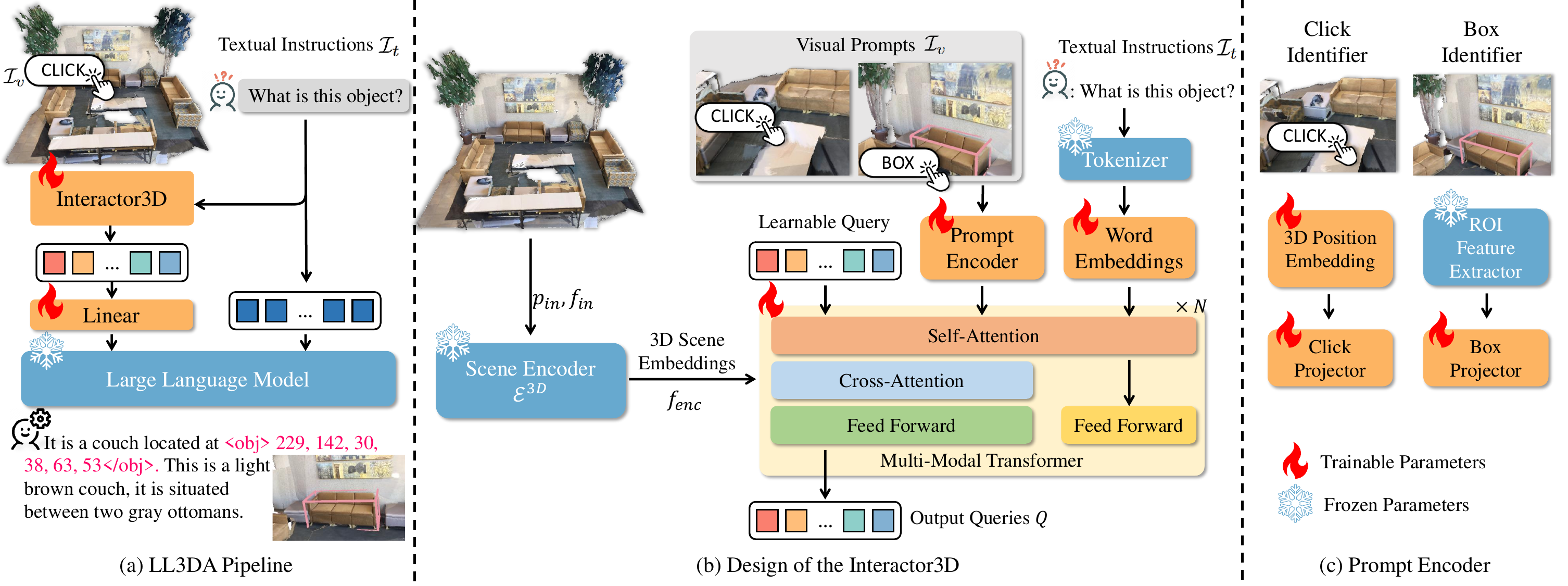}
    \vspace{-5mm}
	\caption{
    	\textbf{Overview of the Proposed Approach.} 
        (a) The overall pipeline of our proposed \modelname{} first extracts interaction-aware 3D scene embeddings, which are later projected to the prefix of textual instructions as the input of a frozen LLM.
        (b) The detailed design of the Interactor3D, which aggregates visual prompts, textual instructions, and 3D scene embeddings into a fixed length querying tokens.
        (c) The prompt encoder encodes the user clicks and box coordinates with the positional embeddings and ROI features, respectively.
    }
	\label{fig:pipeline}
\end{figure*}

\myparagraph{Large Multimodal Models (LMM).}
Along with the rapid development of \textbf{L}arge \textbf{L}anguage \textbf{M}odels (LLM) \cite{zhang2022opt,chung2022scaling}, researchers have made great recent efforts adapting LLMs to visual understanding and reasoning tasks \cite{xu2023survey-multimodal,yin2023survey-mllm}.
Some project or compress global image features as prefix for instruction following \cite{li2023blip2,liu2023llava,ye2023mplug-owl,zhu2023minigpt4}, while others extract ROI features as LLM tokens for region-oriented instruction reasoning \cite{zhang2023gpt4roi,chen2023position}.
Meanwhile, InstructBLIP \cite{instructblip} proposes to extract textual instruction-aware visual features, and has achieved remarkable success addressing complex and unseen instructions.
Concurrently, researchers have also made great attempts solving various 3D tasks using LLMs. 
Notably, \cite{xu2023pointllm,guo2023pointbind+pointllm,luo2023cap3d,zhou2023regionblip} demonstrate remarkable success in understanding and reasoning about 3D objects.
In this paper, we present an LLM-driven solution that could handle both interactions in forms of visual prompts and textual instructions, and propose to extract interaction-aware 3D scene representations for better instruction following.

\section{Methodology}
\label{sec:method}


To build a general purpose agent that could handle both visual and textual interactions in complex 3D environments, we propose \modelname{}, an LLM driven auto-regressive approach to 3D vision language tasks.
In this section, we first introduce the problem formatting in \cref{subsec:problem formatting}.
Next, we introduce our model design in details (\cref{subsec:encoder}).

\subsection{Problem Formatting}
\label{subsec:problem formatting}

\myparagraph{Model I/O.}
As shown in \cref{fig:pipeline} (a), the \textbf{input} of our model consists of a 3D scene represented by a set of points $PC$, the textual instruction $\mathcal{I}_{t}$, and potential visual interactions $\mathcal{I}_{v}$ that serve as supplementary spatial identifiers.
Here, point cloud $PC = \left[p_{in}, f_{in}\right] \in \mathbb{R}^{N \times \left(3 + F\right)}$, where $p_{in} \in \mathbb{R}^{N \times 3}$ and $f_{in} \in \mathbb{R}^{N \times F}$ are the point coordinates and the additional point features, including \textit{color}, \textit{normal}, and \textit{height}.
The \textbf{output} of our model is free-form natural language, part of whom could be interpreted into 3D coordinates.

\myparagraph{Instruction Formatting.}
Following existing LMMs \cite{xu2023pointllm}, we begin the textual instructions $\mathcal{I}_{t}$ with the ``\sep{\textit{\#\#\# human:}}'' identifier, and ask the model to generate responses after the ``\sep{\textit{\#\#\# assistant:}}'' identifier.
This endows the model with the ability to distinguish information from the context, and further engage in multi-turn conversations.

\myparagraph{Coordinate Representations.} To provide LLMs with the capacity to perceive and respond with 3D coordinates, we convert the 3D points and 3D bounding boxes to plain texts.
Specifically, a point is represented by ``\loc{$x$, $y$, $z$}'', and a bounding box is represented by its center point and size, \textit{i.e.} ``\obj{$c_{x}$, $c_{y}$, $c_{z}$, $w$, $h$, $l$}''.
Here, all the numerical data is discretized into unsigned integers within a range of $\left[0, 255\right]$ with respect to the boundary of the input 3D scene.
This design could naturally fit in the vocabulary of existing pre-trained LLMs \cite{zhang2022opt,touvron2023llama}.
Without introducing any additional learnable tokens, we could save the effort of tuning the whole LLM.

\subsection{Model Design}
\label{subsec:encoder}

As shown in \cref{fig:pipeline} (a), our model first aggregates a fixed-length scene embeddings through the Interactor3D, which takes the visual prompts, the textual instructions, and the 3D scene as input.
Next, the aggregated scene embeddings are projected to the prefix of textual instructions as inputs of a frozen LLM.
The detailed design of Interactor3D is shown in \cref{fig:pipeline} (b), which consists of a frozen 3D scene encoder $\mathcal{E}^{3D}$, a visual prompt encoder, and a Q-Former to transform the permutation-invariant 3D embeddings into a fixed-length interaction-aware scene embedding, serving as the prefix of the LLM's input.

\myparagraph{Scene Encoder.} We adopt the masked transformer encoder pre-trained on ScanNet detection \cite{chen2023vote2cap-detr} as the scene encoder, $\mathcal{E}^{3D}$, which takes $PC$ as its input, and outputs the 3D scene embeddings:
\begin{equation}
    f_{enc} = \mathcal{E}^{3D}\left(PC\right) = \mathcal{E}^{3D}\left(p_{in}; f_{in}\right) \in \mathbb{R}^{M \times d}.
\end{equation}
Here, $f_{enc}$ consists of $d$-dimensioned features for $M$ points uniformly down-sampled from the input 3D scene through the \textbf{F}arthest \textbf{P}oint \textbf{S}ampling (FPS) algorithm.
In practice, we choose to keep the scene encoder frozen to save the memory cost during training.

\myparagraph{Visual Prompt Encoder.}
We mainly take two common types of visual interactions into consideration, user clicks and 3D box annotations \cite{kirillov2023segany}.
Each user click is first normalized within a range of $[0, 1]$ by the size of the input 3D scene $p_{\text{click}} \in \mathbb{R}^{3}$.
Then, we encode $p_{\text{click}}$ with the 3D Fourier positional embeddings \cite{tancik2020fourier} function:
\begin{equation}
    \text{pos}\left(p_{\text{click}}\right) = \left[\sin \left(2\pi p_{\text{click}}\cdot B\right); \cos \left(2\pi p_{\text{click}} \cdot B\right)\right].
\end{equation}
Here, $B \in \mathbb{R}^{3 \times \left(d / 2\right)}$ is a learnable matrix.
The box annotation is represented by the ROI feature $f_{\text{box}} \in \mathbb{R}^{d}$ extracted by a pre-trained 3D object detector \cite{chen2023vote2cap-detr}.
The two types of the visual prompts are then projected with separate and identical \textbf{F}eed \textbf{F}orward \textbf{N}etworks (FFN).
\begin{equation}
\begin{aligned}
    f_{\text{click}} &= FFN_{\text{click}}\left(\text{pos}\left(p_{\text{click}}\right)\right) \\
    f_{\text{box}} &= FFN_{\text{box}}\left(f_{\text{box}}\right)
\end{aligned}
\end{equation}
In practice, we represent each visual prompt with 8 tokens.

\myparagraph{Multi-Modal Transformer}(MMT)
serves as a role to 1) address the contradiction between permutation-invariant 3D scene embeddings and the position-sensitive causal LLMs, 2) bridge the gap between frozen unimodal experts, and 3) fill the needs for interaction-aware feature extraction.
Inspired by the Q-Former architecture \cite{instructblip,li2023blip2}, MMT aggregates the visual information within a fixed number of 32 learnable querying tokens.
In each layer, the queries interact with the encoded visual prompts $\left[f_{\text{click}}; f_{\text{box}}\right]$ and the textual instructions $\mathcal{I}_{t}$ through a shared self-attention.
Then, we allow the learnable querying tokens and the visual prompts to interact with the task-agnostic 3D scene embeddings $f_{enc}$ via cross-attention.
The output of MMT is 32 queries noted as $Q \in \mathbb{R}^{32\times 768}$, and are finally projected to the embedding space of LLM with a simple linear projector.
In practice, we notice that initializing Q-Former with pre-trained BERT \cite{devlin2018bert,li2023blip2} weights will lead to repetitive outputs, thus we only choose to initialize the pre-trained word and position embeddings from BERT.

\myparagraph{LLM.}
We consider the decoder-only generative pre-trained transformers \cite{zhang2022opt,touvron2023llama} as our large language model backbone, which are sensitive to the input orders because of the position embeddings, and the causal attention mask.
The parameters and the embedding layers of the LLM are kept frozen to save memory cost.
During inference, we generate the responses via searching for the response $s^{*}$ that satisfies:
\begin{equation}
    s^{*} = \arg \max_{s} P\left(s\vert PC, \mathcal{I}_{t}, \mathcal{I}_{v}\right).
\end{equation}
In practice, we use beam search with a beam size of 4.

\section{Multi-modal Instruction Tuning}

A general purpose agent is meant to deal with various tasks in complex 3D scenes.
Apart from introducing proper training data, it is important to direct the model to generate the desired outputs.
Thus, \cref{subsec:task instructions} will first introduce how we identify each task.
After that, \cref{subsec:training} will present details for training objective.

\subsection{Tasks and Instructions.}
\label{subsec:task instructions}

All the following tasks will be modelled as auto-regressive generation after the ``\sep{\textit{\#\#\# assistant:}}'' identifier.

\myparagraph{3D Dense Captioning} requires the localization and description of instances in diverse 3D environments. 
We adopt either user clicks and box annotations as the visual prompt to identify the object to be described.
Additionally, we design two types of textual instructions that ask the model to either ``describe'' or ``describe and localize'' the object, which diversifies the tasks, and leads to better performance.

\myparagraph{3D Question Answering} requires the model to generate response to the questions based on the global knowledge of a 3D scene.
To help the model better understand the 3D environment, we also design two types of textual instructions that ask the model to either ``answer'' or ``answer and localize the related objects''.
The latter serves as an auxiliary task widely adopted in various 3D-QA methods \cite{azuma2022scanqa,parelli2023clip-guided}.
To diversify the tasks during training, we randomly adopt visual prompts to the objects mentioned in the questions.

\myparagraph{Scene Description} requires the model to translate its global knowledge of the 3D scene into natural languages, thus we simply ask the ``describe'' this 3D scene.

\myparagraph{Embodied Conversation and Planning} could be treated as multi-turn conversation, which we use the ``\sep{\textit{\#\#\# human:}}'' and ``\sep{\textit{\#\#\# assistant:}}'' identifier to distinguish the source of information as introduced in \cref{subsec:problem formatting}.

\subsection{Instruction Following Tuning}
\label{subsec:training}

During training, for tasks that requires additional visual interactions, \textit{i.e.} 3D-DC and 3D-QA, we randomly select the user clicks and box annotations.

\myparagraph{Training Objective.} 
Our training objective is to optimize the trainable parameters $\theta$, so that the likelihood of the target response sequence $s$ is maximized given the input point cloud $PC$, and the interactions $\mathcal{I}_{v}$ and $\mathcal{I}_{t}$:
\begin{equation}
    \theta^{*} = \arg \max_{\theta} P\left(s \vert PC; \mathcal{I}_{v}; \mathcal{I}_{t}; \theta\right).
\end{equation}
In practice, this is accomplished by adopting the token-wise cross-entropy loss that trains the model to predict the $i$th token $s_{\left[i\right]}$ given the previous $(i-1)$ tokens, $s_{\left[1, \cdots, i-1\right]}$.
\begin{equation}
    \mathcal{L}\left(\theta\right) = - \sum_{i=1}^{|s|} \log P\left(s_{\left[i\right]} \vert PC; \mathcal{I}_{v}; \mathcal{I}_{t}; \theta; s_{\left[1, \cdots, i-1\right]}\right).
\end{equation}
Here, $|s|$ is the number of tokens in the desired response.
\section{Experiments}
\label{sec:benchmark}

To examine the capacity of our proposed model, we provide numerous evaluations.
To begin with, we introduce the datasets, metrics, and implementation details (\cref{subsec:exp-intro}).
Then, we compare our model's capacity of understanding and reasoning in complex 3D environments with previous 3D specialists on 3D Dense Captioning and 3D Question Answering (\cref{subsec:exp-sota}), and conduct quantitative ablation studies on the model design and training strategy (\cref{subsec:exp-ablations}).
Finally, \cref{subsec:exp-qualitative} showcases several qualitative results.

\begin{table*}[htbp]
    \caption{
        \textbf{Quantitative Comparisons for 3D Dense Captioning on ScanRefer\cite{chen2020scanrefer} and Nr3D\cite{achlioptas2020referit3d}.}
        For fair comparison, we list methods that are trained under the standard per-word cross-entropy loss without additional 3D scenes.
        We use the box estimations from Vote2Cap-DETR to simulate the box annotations as the visual prompts.
        Our proposed \modelname{} surpasses previous 3D specialists on both datasets.
    }
    \label{tab:benchmark-3d-dense-cap}
    \centering
    \resizebox{\linewidth}{!}{
    \begin{tabular}{c|ccccccccc|cccc}
    \toprule
    \multirow{2}{*}{Method} & \multicolumn{9}{c|}{\large ScanRefer}                                                                                                                            & \multicolumn{4}{c}{\large Nr3D}                                                \\ 
                            & C@0.25$\uparrow$ & B-4@0.25$\uparrow$ & M@0.25$\uparrow$ & R@0.25$\uparrow$ &  & C@0.5$\uparrow$ & B-4@0.5$\uparrow$ & M@0.5$\uparrow$ & R@0.5$\uparrow$ & C@0.5$\uparrow$ & B-4@0.5$\uparrow$ & M@0.5$\uparrow$ & R@0.5$\uparrow$ \\ \hline
    Scan2Cap\cite{chen2021scan2cap}                & 56.82            & 34.18              & 26.29            & 55.27            &  & 39.08           & 23.32             & 21.97           & 44.78           & 27.47           & 17.24             & 21.80           & 49.06           \\
    MORE\cite{jiao2022more}                    & 62.91            & 36.25              & 26.75            & 56.33            &  & 40.94           & 22.93             & 21.66           & 44.42           & -               & -                 & -               & -               \\
    SpaCap3D\cite{wang2022spacap3d}                & -                & -                  & -                & -                &  & 44.02           & 25.26             & 22.33           & 45.36           & 33.71           & 19.92             & 22.61           & 50.50           \\
    REMAN\cite{mao2023REMAN}                   & 62.01            & 36.37              & 26.76            & 56.25            &  & 45.00           & 26.31             & 22.67           & 46.96           & 34.81           & 20.37             & 23.01           & 50.99           \\
    D3Net\cite{chen2021d3net}                   & -                & -                  & -                & -                &  & 46.07           & 30.29             & 24.35           & 51.67           & 33.85           & 20.70             & 23.13           & 53.38           \\
    Contextual\cite{zhong2022contextual3DdenseCap}              & -                & -                  & -                & -                &  & 46.11           & 25.47             & 22.64           & 45.96           & 35.26           & 20.42             & 22.77           & 50.78           \\
    UniT3D\cite{chen2023unit3d}                  & -                & -                  & -                & -                &  & 46.69           & 27.22             & 21.91           & 45.98           & -               & -                 & -               & -               \\
    3DJCG\cite{cai20223djcg}                   & 64.70            & 40.17              & 27.66            & 59.23            &  & 49.48           & 31.03             & 24.22           & 50.80           & 38.06           & 22.82             & 23.77           & 52.99           \\
    3D-VLP\cite{jin2023-3D-VLP}                  & 70.73            & 41.03              & 28.14            & \textbf{59.72}            &  & 54.94           & 32.31             & 24.83           & 51.51           & -               & -                 & -               & -               \\
    3D-VisTA$^{*}$\cite{zhu20233d-vista}                & -                & -                  & -                & -                &  & 61.60           & 34.10             & \textbf{26.80}           & 55.00           & -               & -                 & -               & -               \\
    Vote2Cap-DETR\cite{chen2023vote2cap-detr}           & 71.45            & 39.34              & \textbf{28.25}            & 59.33            &  & 61.81           & 34.46             & 26.22           & 54.40           & 43.84           & 26.68             & 25.41           & 54.43           \\
    %
    %
    \modelname{} (Ours)            & \textbf{74.17}            & \textbf{41.41}              & 27.76            & 59.53            &  & \textbf{65.19}            & \textbf{36.79}          & 25.97           & \textbf{55.06}           & \textbf{51.18}           & \textbf{28.75}             & \textbf{25.91}           & \textbf{56.61}           \\
    \bottomrule
    \end{tabular}
    }
    \label{exp:comparison_on_scanrefer}
\end{table*}
\begin{table*}[htbp]
    \caption{
        \textbf{Quantitative Comparisons for 3D Question Answering on ScanQA\cite{azuma2022scanqa}.}
        We categorize previous works into classification based (``CLS'') and generation based (``GEN'') methods.
        The results from 3D-LLM$^*$ come from their fine-tuned version.
        \modelname{} out-performs previous methods on the validation set and two test sets.
    }
    \label{tab:benchmark-scanqa}
    \centering
    \resizebox{0.85\linewidth}{!}{
    \begin{tabular}{cccccccccccccccc}
    \toprule
    \multirow{2}{*}{Method} & \multirow{2}{*}{\begin{tabular}[c]{@{}c@{}}Answer\\ Type\end{tabular}} & \multicolumn{4}{c}{Validation}                          &  & \multicolumn{4}{c}{Test w/ object}                      &  & \multicolumn{4}{c}{Test w/o object}                     \\ \cline{3-6} \cline{8-11} \cline{13-16} 
                            &                                                                        & C$\uparrow$ & B-4$\uparrow$ & M$\uparrow$ & R$\uparrow$ &  & C$\uparrow$ & B-4$\uparrow$ & M$\uparrow$ & R$\uparrow$ &  & C$\uparrow$ & B-4$\uparrow$ & M$\uparrow$ & R$\uparrow$ \\ \hline
    ScanQA\cite{azuma2022scanqa}                  & \multirow{6}{*}{CLS}                                                   & 64.86       & 10.08         & 13.14       & 33.33       &  & 67.29       & 12.04         & 13.55       & 34.34       &  & 60.24       & 10.75         & 12.59       & 31.09       \\
    Clip-Guided\cite{parelli2023clip-guided}             &                                                                        & -           & -             & -           & -           &  & 69.53       & \textbf{14.64}         & 13.94       & 35.15       &  & 62.83       & 11.73         & 13.28       & 32.41       \\
    Multi-CLIP\cite{delitzas2023multi-clip}              &                                                                        & -           & -             & -           & -           &  & 68.70       & 12.65         & 13.97       & 35.46       &  & 63.20       & 12.87         & 13.36       & 32.61       \\
    3D-VLP\cite{jin2023-3D-VLP}                  &                                                                        & 66.97       & 11.15         & 13.53       & 34.51       &  & 70.18       & 11.23         & 14.16       & 35.97       &  & 63.40       & \textbf{15.84}         & 13.13       & 31.79       \\
    3D-VisTA\cite{zhu20233d-vista}                &                                                                        & -           & -             & -           & -           &  & 68.60       & 10.50         & 13.80       & 35.50       &  & 55.70       & 8.70          & 11.69       & 29.60       \\
    \hline
    3D-LLM$^*$\cite{hong20233d-llm}                  & \multirow{2}{*}{GEN}                                                   & 69.40       & 12.00         & 14.50       & 35.70       &  & 69.60       & 11.60         & 14.90       & 35.30       &  & -           & -             & -           & -           \\ 
    %
    \modelname{} (Ours)             &                                                                        & \textbf{76.79}       & \textbf{13.53}         & \textbf{15.88}       & \textbf{37.31}       &  & \textbf{78.16}       & 13.97         & \textbf{16.38}       & \textbf{38.15}       &  & \textbf{70.29}       & 12.19         & \textbf{14.85}       & \textbf{35.17}       \\
    %
    \bottomrule
    \end{tabular}
    }
\end{table*}

\subsection{Datasets, Metrics and Implementation Details}
\label{subsec:exp-intro}
\myparagraph{Datasets.} In this paper, we experiment with 3D data from ScanNet \cite{dai2017scannet}, a 3D dataset covering 1,201 and 312 diverse and complex indoor 3D scenes for training and validation.
The language annotations used in this study are sourced from ScanRefer \cite{chen2020scanrefer}, Nr3D \cite{achlioptas2020referit3d}, ScanQA \cite{azuma2022scanqa}, and the ScanNet subset of 3D-LLM \cite{hong20233d-llm}.
This combination covers a variety of tasks, including instance and scene descriptions, conversations, embodied planning and question answering.
Please refer to the supplementary materials for more details on the statistics of data.

\myparagraph{Metrics.} Here, we adopt \textbf{C}, \textbf{B-4}, \textbf{M}, \textbf{R} as abbreviations for CiDEr \cite{vedantam2015cider}, BLEU-4 \cite{papineni2002bleu}, METEOR \cite{banerjee2005meteor}, and Rouge-L \cite{lin2004rouge} to evaluate the quality of the generated textual responses.

\myparagraph{Implementation Details.} Following previous works on 3D vision language tasks \cite{chen2023vote2cap-detr,chen2021scan2cap}, we randomly sample $40$k points from each 3D scene as the 3D input.
We adopt the pre-trained OPT-1.3B \cite{zhang2022opt} as our causal LLM backbone, which is frozen and loaded in float16 to save memory cost.
We adopt the AdamW \cite{loshchilov2017AdamW} optimizer with a weight decay of $0.1$ and a learning rate decaying from $10^{-4}$ to $10^{-6}$ with a cosine annealing scheduler for about $100$k iterations.
For all the training tasks, we train with a total batch size of 16, and evaluate our method every 4k iterations.
Each training process consumes no more than eight Nvidia RTX3090 (24G) GPUs within a day.

\subsection{Comparison with SoTA Specialists}
\label{subsec:exp-sota}

We evaluate the model's capacity to understand and reason in 3D environments via 3D-DC and 3D-QA.
For each evaluation task, we fine-tune the trainable parameters in our model on each task for $\sim$30k iterations.

\myparagraph{3D Dense Captioning} demands a model to localize and describe any instance in a 3D scene. 
We benchmarks state-of-the-art methods on the widely-used ScanRefer \cite{chen2020scanrefer} and Nr3D \cite{achlioptas2020referit3d} dataset in \cref{tab:benchmark-3d-dense-cap} under the $m@k$IoU metric \cite{chen2021scan2cap}.
%
%
Here, $m\in \left\{\text{C, B-4, M, R}\right\}$, and the $m$ score of a caption is set to $0$ if the IoU between the predicted box and the object is less than the given threshold $k$.
Following existing works \cite{chen2021scan2cap,chen2023vote2cap-detr}, we consider C@0.25 and C@0.5 as the main metric for ScanRefer, and C@0.5 for Nr3D.
Among the listed methods, UniT3D \cite{chen2023unit3d}, 3DJCG \cite{cai20223djcg}, and 3D-VLP \cite{jin2023-3D-VLP} are pre-trained on multiple 3D vision and language tasks annotated on ScanNet scenes.
Additionally, UniT3D \cite{chen2023unit3d} adopts off-the-shelf image caption models \cite{mokady2021clipcap} and multi-view images to generate additional instance-captions for pre-training.
It is worth mentioning that the results of 3D-VisTA \cite{zhu20233d-vista} come from their reported version that is not trained on additional 3D scenes.
To evaluate our model, we adopt the box predictions produced by Vote2Cap-DETR \cite{chen2023vote2cap-detr} as the visual prompt.
Results show that our method consistently outperforms existing methods on all both datasets.
For example, our method achieves 65.19\% C@0.5 on ScanRefer and 51.18\% C@0.5 on Nr3D, which is (\better{+3.38}\% and \better{+7.34}\%) higher than the current state-of-the-art 3D vision and language model, Vote2Cap-DETR.

\myparagraph{3D Question Answering} requires a model to generate responses to the natural language queries questioning towards an 3D scene.
We benchmark state-of-the-art methods on the ScanQA \cite{azuma2022scanqa} validation set as well as two test benchmarks in \cref{tab:benchmark-scanqa}, and consider CiDEr as the main metric.
The majority of the listed methods are based on classification (marked ``CLS''), \textit{i.e.}, selecting responses from a pre-defined answer set.
Meanwhile, 3D-LLM \cite{hong20233d-llm} tries to address 3D-QA via generating texts (marked ``GEN''), and we list their fine-tuned version for comparison.
Results show that our method consistently outperforms existing methods on all the evaluation sets, and surpasses the generation based method, 3D-LLM, by a large margin (\better{+7.39}\% CiDEr score on the validation set).

\subsection{Ablation Studies}
\label{subsec:exp-ablations}

In this section, we provide ablation studies on model designs and training strategies. 
We evaluate on ScanRefer and ScanQA to quantize the effectiveness.

\begin{figure}[htbp]
	\centering
	\includegraphics[width=\linewidth]{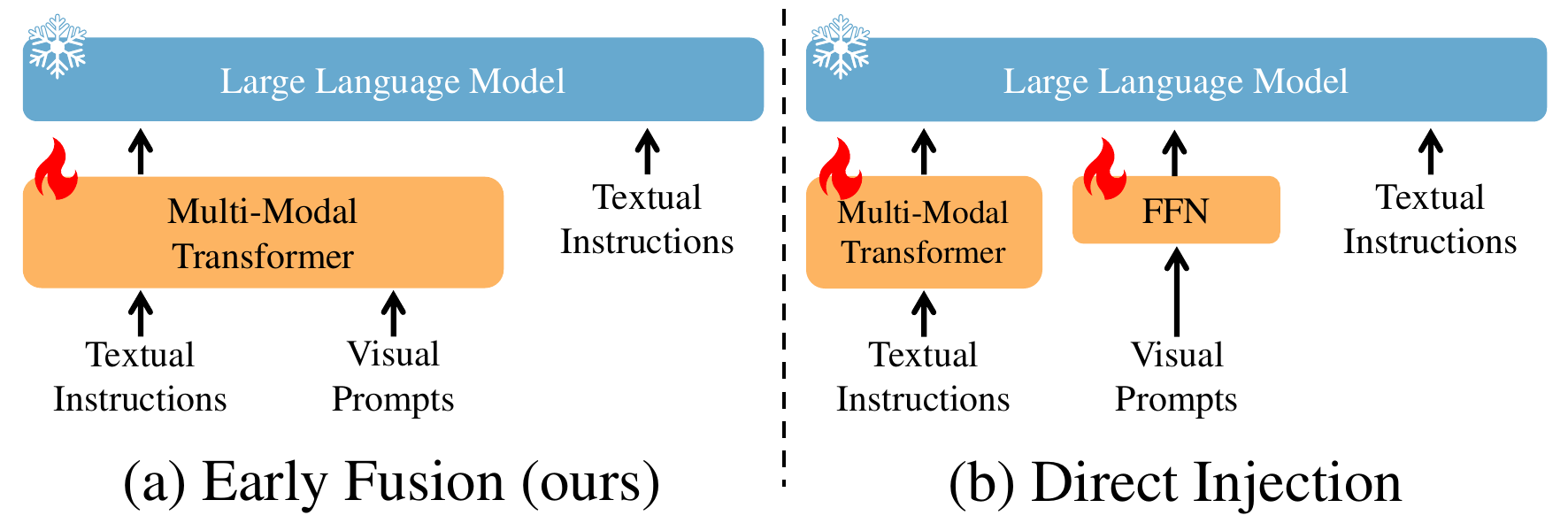}
    \vspace{-5mm}
	\caption{
    	\textbf{Different Ways of Encoding Visual Prompts.} 
        We listed two ways of encoding visual prompts, (a) adopting a unified transformer to aggregate features from all kinds of interactions, and (b) directly concatenate the visual prompts to the scene embeddings.
        Experiments (\cref{tab:ablation-qformer-design}) show that early fusion(a) leads to a better performance.
    }
	\label{fig:ablation-qformer}
\end{figure}

\myparagraph{Effectiveness of the Q-Former Design.}
We list two ways of utilizing visual prompts in \cref{fig:ablation-qformer},
where (a) is our proposed method that adopts a unified transformer that aggregates information from both the textual instructions and visual prompts, and (b) is the ``direct injection'' version, which only extract instruction-aware 3D feature with visual prompts concatenated after the scene embeddings.
We train both models from scratch and evaluate their performance on ScanRefer 3D Dense Captioning.
The results (\cref{fig:ablation-qformer}) show that the method we use (\cref{fig:ablation-qformer} (a)) could better capture feature related to the visual prompts, leading to better instance caption generation performance (\better{+3.45\%} C@0.5).

\begin{table}[htbp]
\caption{
        \textbf{Effectiveness of Q-Former Design on ScanRefer\cite{chen2020scanrefer}.}
        We design two different ways of utilizing visual prompts.
        The ``early fusion'' enables direct interaction with the 3D scene, thus it achieves a better performance.
    }
    \label{tab:ablation-qformer-design}
    \centering
    \resizebox{\linewidth}{!}{
    \begin{tabular}{ccccc}
    \toprule
    Visual Prompt  & C@0.5$\uparrow$ & B-4@0.5$\uparrow$ & M@0.5$\uparrow$ & R@0.5$\uparrow$ \\ \hline
    direct         & 59.39           & 33.27             & 25.19           & 53.39           \\
    ours           & \textbf{62.84}  & \textbf{35.81}    & \textbf{25.81}  & \textbf{54.45}  \\ 
    \bottomrule
    \end{tabular}
}
\end{table}

\myparagraph{Instructions as Auxiliary Tasks for 3D Dense Captioning.} 
We have introduced two types of textual instructions in \cref{subsec:task instructions} for 3D-DC, \textit{i.e.} the ``describe'' only instructions and ``detect and localize'' instructions.
Additionally, we have introduced two types of visual prompts (\cref{fig:pipeline} \& \cref{subsec:training}).
In this study, we show how they serve as auxiliary tasks for 3D-DC by evaluating on ScanRefer in \cref{tab:ablation-densecap-auxiliary}.
All the methods listed are trained from scratch.
In \cref{tab:ablation-densecap-auxiliary}, ``Aux.Loc'' identifies whether we train the model with the ``detect and localize'' instructions, and ``Clicks'' identifies whether we train the model with clicks as additional visual prompts.
Results show that either way can serve as good auxiliary tasks for 3D Dense Captioning.

\begin{table}[htbp]
\caption{
        \textbf{Effectiveness of Instructions as 3D Dense Captioning Auxiliary Tasks.}
        We train the models from scratch and evaluate on ScanRefer\cite{chen2020scanrefer}.
        ``Aux.Loc'' identifies whether we train with the ``describe and localize'' instructions.
        ``Clicks'' identifies whether we train with clicks as additional visual prompts.
    }
    \label{tab:ablation-densecap-auxiliary}
    \centering
    \resizebox{\linewidth}{!}{
    \begin{tabular}{cccccc}
    \toprule
    Aux.Loc      & Clicks       & C@0.5$\uparrow$ & B-4@0.5$\uparrow$ & M@0.5$\uparrow$ & R@0.5$\uparrow$ \\ \hline
    -            & -            & 60.85           & 34.09             & 25.53           & 53.48           \\
    $\checkmark$ & -            & 61.81           & 34.15             & 25.49           & 53.83           \\
    -            & $\checkmark$ & 62.20           & 34.26             & 25.67           & 53.87           \\
    %
    $\checkmark$ & $\checkmark$ & \textbf{62.84}  & \textbf{35.81}    & \textbf{25.81}  & \textbf{54.45}  \\ 
    \bottomrule
    \end{tabular}
    }
\end{table}

\begin{table*}[htbp]
    \caption{
        \textbf{Evaluation as a Generalist.}
        The first three rows list the performance of models trained from scratch as experts on each dataset.
        The results in the following three rows belong to the model fine-tuned from the generalist weights.
        The last row evaluates the model trained as a generalist.
        ScanRefer\cite{chen2020scanrefer} and Nr3D\cite{achlioptas2020referit3d} are used to evaluate the dense captioning performance, and ScanQA\cite{azuma2022scanqa} is used to evaluate the question answering performance.
        Serving as a generalist, our method can differentiate each task, and produce strong results based on textual instructions and visual prompts.
    }
    \label{tab:ablation-generalist}
    \centering
    \resizebox{\linewidth}{!}{
    \begin{tabular}{ccccccccccccccc}
    \toprule
    \multirow{2}{*}{Method} & \multicolumn{4}{c}{ScanRefer}                                           &  & \multicolumn{4}{c}{Nr3D}                                                 &  & \multicolumn{4}{c}{ScanQA}\\ \cline{2-5} \cline{7-10} \cline{12-15} 
                            & C@0.5$\uparrow$ & B-4@0.5$\uparrow$ & M@0.5$\uparrow$ & R@0.5$\uparrow$ &  & C@0.5$\uparrow$ & B-4@0.5$\uparrow$ & M@0.5$\uparrow$ & R@0.5$\uparrow$  &  & C$\uparrow$     & B-4 $\uparrow$     & M$\uparrow$     & R$\uparrow$    \\ \hline
    ScanRefer(scratch)      & 62.84           & 35.81             & 25.81           & 54.45           &  & -               & -                 & -              & -                &  & -               & -                          & -               & -    \\
    Nr3D(scratch)           & -               & -                 & -               & -               &  & 44.95           & 27.67             & 25.67     
        & 55.79            &  & -               & -                          & -               & -    \\
    ScanQA(scratch)         & -               & -                 & -               & -               &  & -               & -                  & -              & -                &  & 74.80           & \textbf{13.68}             & 15.40           & 36.25\\
    \cline{1-1}
    ScanRefer(fine-tuned)   & \textbf{65.19}  & \textbf{36.79}    & \textbf{25.97}  & \textbf{55.06}  &  & -               & -                 & -              & -                &  & -               & -                          & -               & -    \\
    Nr3D(fine-tuned)        & -               & -                 & -               & -               &  &  \textbf{51.18} & \textbf{28.75}    & \textbf{25.91}           & \textbf{56.61}            &  & -   & -   & -   & -  \\ 
    ScanQA(fine-tuned)      & -               & -                 & -               & -               &  & -               & -                  & -               & -                &  & \textbf{76.79}  & 13.53            & \textbf{15.88}  & \textbf{37.31}   \\ 
    \cline{1-1}
    w/o fine-tuning         & 62.98           & 35.97             & 25.66           & 54.65           &  & 23.94           & 13.37              & 22.31          & 45.78            &  & 75.67           & 13.33             & 15.37           & 37.02\\
    \bottomrule
    \end{tabular}
    }
\end{table*}

\begin{table}[htbp]
\caption{
        \textbf{Effectiveness of Interactions as 3D Question Answering Auxiliary Tasks.}
        We train the model from scratch and evaluate all the models from scratch on ScanQA\cite{azuma2022scanqa} validation set.
        ``Aux.Loc'' identifies whether we train with the ``answer and localize'' instructions, and ``Visual Prompts'' identifies whether we train with visual prompts.
    }
    \label{tab:ablation-scanqa-auxiliary}
    \centering
    \resizebox{\linewidth}{!}{
    \begin{tabular}{cccccc}
    \toprule
    Aux.Loc      & Visual Prompts & CiDEr$\uparrow$ & BLEU-4$\uparrow$  & METEOR$\uparrow$ & Rouge-L$\uparrow$ \\ \hline
    -            & -              & 67.85           & 11.87             & 13.96            & 33.87           \\
    $\checkmark$ & -              & 72.73           & 13.27             & 14.90            & 35.87           \\
    -            & $\checkmark$   & 68.09           & 12.59             & 14.20            & 33.71           \\
    $\checkmark$ & $\checkmark$   & \textbf{74.80}  & \textbf{13.68}    & \textbf{15.40}   & \textbf{36.25}  \\ 
    \bottomrule
    \end{tabular}
    }
\end{table}

\myparagraph{Instructions as Auxiliary Tasks for 3D Question Answering.} 
We have made similar study on the effectiveness of treating ``answer and localize'' instructions and additional visual prompts as auxiliary tasks for 3D-QA on ScanQA \cite{azuma2022scanqa} validation set in \cref{tab:ablation-scanqa-auxiliary}.
The listed methods are evaluated without any visual interactions for fair comparison.
Results show that the additional textual instructions and visual prompts improve the task diversity and further improve the performance on 3D Question Answering.

\myparagraph{Performance as a Generalist.}
To examine whether our method could distinguish different tasks given the textual instructions and visual prompts introduced in \cref{subsec:task instructions}, we evaluate our model on different tasks with the same set of weights in \cref{tab:ablation-generalist} with several baseline methods.
The first three rows list the performance our proposed models trained on each dataset from scratch as generalists.
The following three rows list the performance of the model fine-tuned on each dataset.
The last row lists the performance of our model as a generalist.
Results show that our model could distinguish 3D-DC and 3D-QA given the textual instructions and visual prompts, providing strong results (62.98\% C@0.5 on ScanRefer, 75.67\% CiDEr on ScanQA).
However, the generalist model achieves poor results on Nr3D \cite{achlioptas2020referit3d}, which is because we did not try to differentiate between Nr3D and ScanRefer during training as these two datasets are used for the same task in the first place.
There is also an interesting observation that although we did not differentiate between the two datasets for 3D-DC, and the training sample sizes of the two models were similar, the model still tend to achieve high scores on ScanRefer (62.98\% C@0.5).
We are also excited to see that the weights of the generalist model could serve as a strong initialization of weights for fine-tuning.
For example, the fine-tuned model on ScanRefer could achieve 65.19\% C@0.5, which is \better{+2.35}\% higher than the model trained from scratch.

\myparagraph{Importance of Textual Instructions.}
We further conduct study to see whether the instructions are necessary in 3D Dense Captioning in \cref{tab:ablation-interaction}.
The first row is our baseline method that directly generates the captions based on visual prompts without any textual instructions, and the second row is our method that is trained with textual instructions introduced in \cref{subsec:task instructions}.
Both methods are trained from scratch for fair comparison.
We notice that since the LLM is frozen, certain textual instructions are beneficial to generating results in specific domains/tasks.

\begin{table}[htbp]
    \caption{
        \textbf{Effectiveness of Instructions on 3D Dense Captioning.}
        We perform experiments on ScanRefer\cite{chen2020scanrefer}.
        The baseline method directly generates the captions given the input 3D scene and visual prompts without any textual instructions.
    }
    \label{tab:ablation-interaction}
    \centering
    \resizebox{\linewidth}{!}{
    \begin{tabular}{ccccc}
    \toprule
    Instructions   & C@0.5$\uparrow$ & B-4@0.5$\uparrow$ & M@0.5$\uparrow$ & R@0.5$\uparrow$ \\ \hline
    -              & 60.20           & 34.79             & 25.40           & 54.03           \\
    $\checkmark$   & \textbf{62.84}  & \textbf{35.81}    & \textbf{25.81}  & \textbf{54.45}  \\ 
    \bottomrule
    \end{tabular}
    }
\end{table}

\myparagraph{Clicks for Better Question Answering.}
One major challenge of answering questions in complex 3D environments is the vague identification of objects with plain texts.
Therefore, we try to click on some of the related objects along with the textual instructions during evaluation, and see how it could affect the generated answers on the ScanQA validation set in \cref{tab:ablation-click-scanqa}.
Results show that this technique would remove the ambiguities, and further improve the quality of the answers (\better{+6.12}\% C).
This illustrates the importance of visual interaction in complex 3D environments.

\begin{table}[htbp]
\caption{
        \textbf{Test Time Visual Interactions for Question Answering on ScanQA\cite{azuma2022scanqa}.}
        The model achieves better performance on the question answering when we add visual prompts to some of the related objects along with the text instructions during evaluation.
    }
    \label{tab:ablation-click-scanqa}
    \centering
    \resizebox{\linewidth}{!}{
        \begin{tabular}{ccccc}
        \toprule
        Visual Prompts  & CiDEr$\uparrow$    & BLEU-4$\uparrow$  & METEOR$\uparrow$    & Rouge-L$\uparrow$    \\ \hline
        -               & 76.79          & \textbf{13.53} & 15.88          & 37.31          \\
        $\checkmark$    & \textbf{82.91} & 11.80          & \textbf{16.74} & \textbf{39.97} \\ 
        \bottomrule
        \end{tabular}
    }
\end{table}

\begin{figure*}[htbp]
	\centering
	\includegraphics[width=\linewidth]{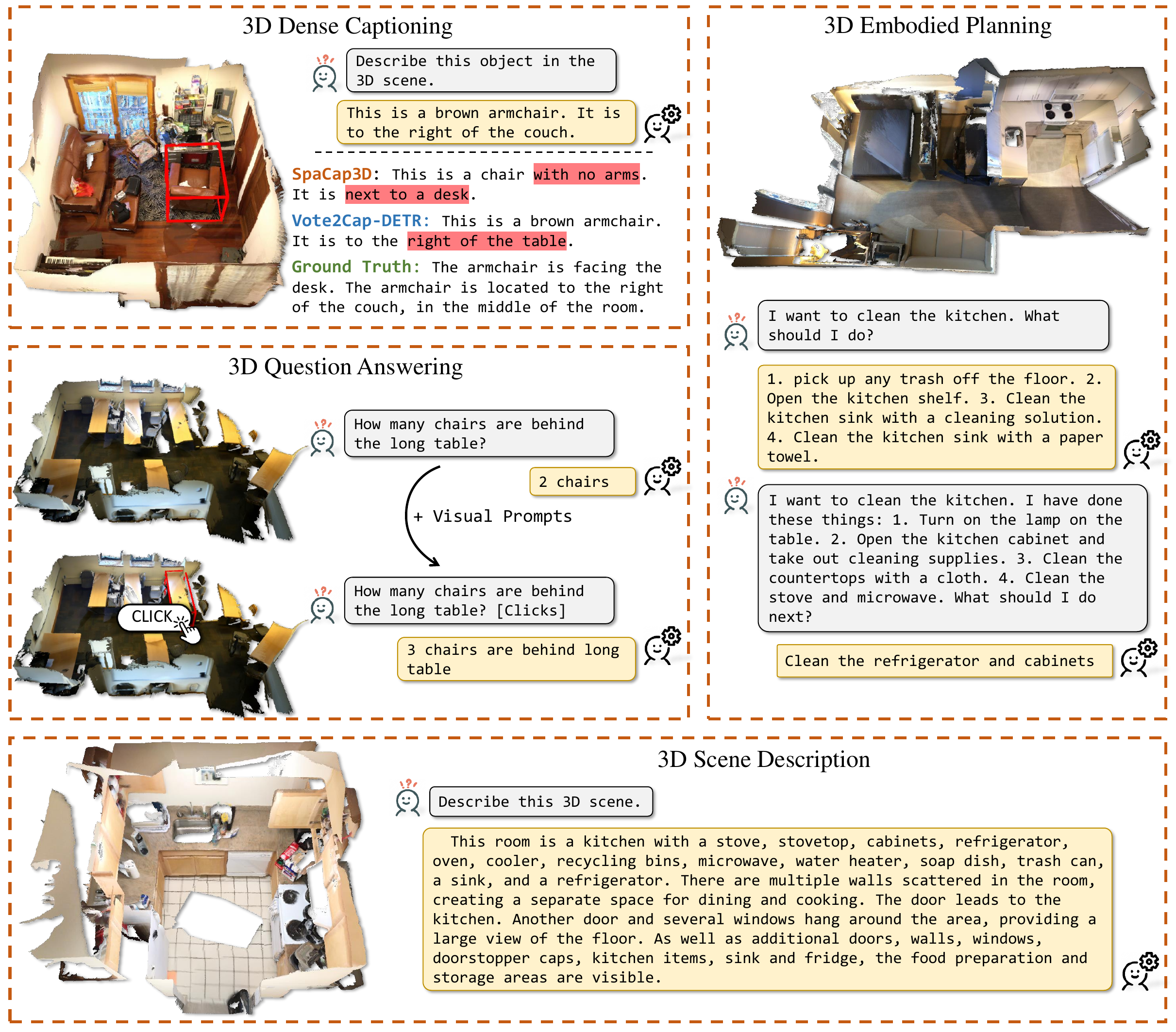}
    \vspace{-5mm}
	\caption{
    	\textbf{Qualitative Results.} 
        We provide several visualization results on various 3D vision and language tasks in diverse 3D environments (living room, classroom, kitchen, and bedroom).
        \colorbox[RGB]{255, 124, 128}{Red} highlights the wrong answer. 
    }
	\label{fig:qualitative}
\end{figure*}

\subsection{Qualitative Results}
\label{subsec:exp-qualitative}
We present several visualization results on different tasks in \cref{fig:qualitative} to show our model's capacities in understanding, reasoning, and planning in different 3D environments.
To prevent repetition when generating long sequences, we combine the top-$k$ \cite{fan2018topk} and top-$p$ \cite{holtzman2019topp} sampling strategy with $k=50$ and $p=0.95$.

\section{Conclusions}

In this paper, we present \modelname{}, a large language 3D assistant that could take both textual- and visual- interactions from human for understanding, reasoning, and planning in complex 3D environments.
Our model directly encodes 3D point cloud for scene representations, and aggregates information from scenes and human interactions with the attention mechanism.
The visual interactions could remove the ambiguities in cluttered 3D environments, showing mighty instruction-following capacities.
Experimental results show that our method could achieve remarkable results on various 3D vision-language benchmarks.
We hope that our approach could inspire further designs and training strategies for large 3D language models. 
In future studies, we anticipate that the availability of high-quality and diverse 3D vision and language annotations will enhance the model's reasoning and planning capabilities.

{\small
\bibliographystyle{ieeenat_fullname}
\bibliography{reference}

\begin{thebibliography}{65}
\providecommand{\natexlab}[1]{#1}
\providecommand{\url}[1]{\texttt{#1}}
\expandafter\ifx\csname urlstyle\endcsname\relax
  \providecommand{\doi}[1]{doi: #1}\else
  \providecommand{\doi}{doi: \begingroup \urlstyle{rm}\Url}\fi

\bibitem[Achlioptas et~al.(2020)Achlioptas, Abdelreheem, Xia, Elhoseiny, and Guibas]{achlioptas2020referit3d}
Panos Achlioptas, Ahmed Abdelreheem, Fei Xia, Mohamed Elhoseiny, and Leonidas Guibas.
\newblock Referit3d: Neural listeners for fine-grained 3d object identification in real-world scenes.
\newblock In \emph{European Conference on Computer Vision}, pages 422--440. Springer, 2020.

\bibitem[Azuma et~al.(2022)Azuma, Miyanishi, Kurita, and Kawanabe]{azuma2022scanqa}
Daichi Azuma, Taiki Miyanishi, Shuhei Kurita, and Motoaki Kawanabe.
\newblock Scanqa: 3d question answering for spatial scene understanding.
\newblock In \emph{proceedings of the IEEE/CVF conference on computer vision and pattern recognition}, pages 19129--19139, 2022.

\bibitem[Banerjee and Lavie(2005)]{banerjee2005meteor}
Satanjeev Banerjee and Alon Lavie.
\newblock Meteor: An automatic metric for mt evaluation with improved correlation with human judgments.
\newblock In \emph{Proceedings of the acl workshop on intrinsic and extrinsic evaluation measures for machine translation and/or summarization}, pages 65--72, 2005.

\bibitem[Cai et~al.(2022)Cai, Zhao, Zhang, Sheng, and Xu]{cai20223djcg}
Daigang Cai, Lichen Zhao, Jing Zhang, Lu Sheng, and Dong Xu.
\newblock 3djcg: A unified framework for joint dense captioning and visual grounding on 3d point clouds.
\newblock In \emph{Proceedings of the IEEE/CVF Conference on Computer Vision and Pattern Recognition}, pages 16464--16473, 2022.

\bibitem[Chen et~al.(2023{\natexlab{a}})Chen, Qin, Luo, Mi, Li, Sun, and Liu]{chen2023position}
Chi Chen, Ruoyu Qin, Fuwen Luo, Xiaoyue Mi, Peng Li, Maosong Sun, and Yang Liu.
\newblock Position-enhanced visual instruction tuning for multimodal large language models.
\newblock \emph{arXiv preprint arXiv:2308.13437}, 2023{\natexlab{a}}.

\bibitem[Chen et~al.(2020)Chen, Chang, and Nie{\ss}ner]{chen2020scanrefer}
Dave~Zhenyu Chen, Angel~X Chang, and Matthias Nie{\ss}ner.
\newblock Scanrefer: 3d object localization in rgb-d scans using natural language.
\newblock In \emph{European Conference on Computer Vision}, pages 202--221. Springer, 2020.

\bibitem[Chen et~al.(2021{\natexlab{a}})Chen, Wu, Nie{\ss}ner, and Chang]{chen2021d3net}
Dave~Zhenyu Chen, Qirui Wu, Matthias Nie{\ss}ner, and Angel~X Chang.
\newblock D3net: A speaker-listener architecture for semi-supervised dense captioning and visual grounding in rgb-d scans.
\newblock \emph{arXiv preprint arXiv:2112.01551}, 2021{\natexlab{a}}.

\bibitem[Chen et~al.(2023{\natexlab{b}})Chen, Sinavski, H{\"u}nermann, Karnsund, Willmott, Birch, Maund, and Shotton]{chen2023driving}
Long Chen, Oleg Sinavski, Jan H{\"u}nermann, Alice Karnsund, Andrew~James Willmott, Danny Birch, Daniel Maund, and Jamie Shotton.
\newblock Driving with llms: Fusing object-level vector modality for explainable autonomous driving.
\newblock \emph{arXiv preprint arXiv:2310.01957}, 2023{\natexlab{b}}.

\bibitem[Chen et~al.(2023{\natexlab{c}})Chen, Zhu, Chen, Lei, Yu, and Chen]{chen2023vote2cap-detr}
Sijin Chen, Hongyuan Zhu, Xin Chen, Yinjie Lei, Gang Yu, and Tao Chen.
\newblock End-to-end 3d dense captioning with vote2cap-detr.
\newblock In \emph{Proceedings of the IEEE/CVF Conference on Computer Vision and Pattern Recognition}, pages 11124--11133, 2023{\natexlab{c}}.

\bibitem[Chen et~al.(2023{\natexlab{d}})Chen, Zhu, Li, Chen, Guo, Lei, Yu, Li, and Chen]{chen2023vote2cap-detr++}
Sijin Chen, Hongyuan Zhu, Mingsheng Li, Xin Chen, Peng Guo, Yinjie Lei, Gang Yu, Taihao Li, and Tao Chen.
\newblock Vote2cap-detr++: Decoupling localization and describing for end-to-end 3d dense captioning.
\newblock \emph{arXiv preprint arXiv:2309.02999}, 2023{\natexlab{d}}.

\bibitem[Chen et~al.(2021{\natexlab{b}})Chen, Gholami, Nie{\ss}ner, and Chang]{chen2021scan2cap}
Zhenyu Chen, Ali Gholami, Matthias Nie{\ss}ner, and Angel~X Chang.
\newblock Scan2cap: Context-aware dense captioning in rgb-d scans.
\newblock In \emph{Proceedings of the IEEE/CVF Conference on Computer Vision and Pattern Recognition}, pages 3193--3203, 2021{\natexlab{b}}.

\bibitem[Chen et~al.(2023{\natexlab{e}})Chen, Hu, Chen, Nie{\ss}ner, and Chang]{chen2023unit3d}
Zhenyu Chen, Ronghang Hu, Xinlei Chen, Matthias Nie{\ss}ner, and Angel~X Chang.
\newblock Unit3d: A unified transformer for 3d dense captioning and visual grounding.
\newblock In \emph{Proceedings of the IEEE/CVF International Conference on Computer Vision}, pages 18109--18119, 2023{\natexlab{e}}.

\bibitem[Chung et~al.(2022{\natexlab{a}})Chung, Hou, Longpre, Zoph, Tay, Fedus, Li, Wang, Dehghani, Brahma, et~al.]{chung2022flan-t5}
Hyung~Won Chung, Le Hou, Shayne Longpre, Barret Zoph, Yi Tay, William Fedus, Eric Li, Xuezhi Wang, Mostafa Dehghani, Siddhartha Brahma, et~al.
\newblock Scaling instruction-finetuned language models.
\newblock \emph{arXiv preprint arXiv:2210.11416}, 2022{\natexlab{a}}.

\bibitem[Chung et~al.(2022{\natexlab{b}})Chung, Hou, Longpre, Zoph, Tay, Fedus, Li, Wang, Dehghani, Brahma, et~al.]{chung2022scaling}
Hyung~Won Chung, Le Hou, Shayne Longpre, Barret Zoph, Yi Tay, William Fedus, Yunxuan Li, Xuezhi Wang, Mostafa Dehghani, Siddhartha Brahma, et~al.
\newblock Scaling instruction-finetuned language models.
\newblock \emph{arXiv preprint arXiv:2210.11416}, 2022{\natexlab{b}}.

\bibitem[Dai et~al.(2017)Dai, Chang, Savva, Halber, Funkhouser, and Nie{\ss}ner]{dai2017scannet}
Angela Dai, Angel~X Chang, Manolis Savva, Maciej Halber, Thomas Funkhouser, and Matthias Nie{\ss}ner.
\newblock Scannet: Richly-annotated 3d reconstructions of indoor scenes.
\newblock In \emph{Proceedings of the IEEE conference on computer vision and pattern recognition}, pages 5828--5839, 2017.

\bibitem[Dai et~al.(2023)Dai, Li, Li, Tiong, Zhao, Wang, Li, Fung, and Hoi]{instructblip}
Wenliang Dai, Junnan Li, Dongxu Li, Anthony Meng~Huat Tiong, Junqi Zhao, Weisheng Wang, Boyang Li, Pascale Fung, and Steven Hoi.
\newblock Instructblip: Towards general-purpose vision-language models with instruction tuning, 2023.

\bibitem[Delitzas et~al.(2023)Delitzas, Parelli, Hars, Vlassis, Anagnostidis, Bachmann, and Hofmann]{delitzas2023multi-clip}
Alexandros Delitzas, Maria Parelli, Nikolas Hars, Georgios Vlassis, Sotirios Anagnostidis, Gregor Bachmann, and Thomas Hofmann.
\newblock Multi-clip: Contrastive vision-language pre-training for question answering tasks in 3d scenes.
\newblock \emph{arXiv preprint arXiv:2306.02329}, 2023.

\bibitem[Devlin et~al.(2018)Devlin, Chang, Lee, and Toutanova]{devlin2018bert}
Jacob Devlin, Ming-Wei Chang, Kenton Lee, and Kristina Toutanova.
\newblock Bert: Pre-training of deep bidirectional transformers for language understanding.
\newblock \emph{arXiv preprint arXiv:1810.04805}, 2018.

\bibitem[Ding et~al.(2023)Ding, Yang, Xue, Zhang, Bai, and Qi]{ding2023pla}
Runyu Ding, Jihan Yang, Chuhui Xue, Wenqing Zhang, Song Bai, and Xiaojuan Qi.
\newblock Pla: Language-driven open-vocabulary 3d scene understanding.
\newblock In \emph{Proceedings of the IEEE/CVF Conference on Computer Vision and Pattern Recognition}, pages 7010--7019, 2023.

\bibitem[Driess et~al.(2023)Driess, Xia, Sajjadi, Lynch, Chowdhery, Ichter, Wahid, Tompson, Vuong, Yu, et~al.]{driess2023palm}
Danny Driess, Fei Xia, Mehdi~SM Sajjadi, Corey Lynch, Aakanksha Chowdhery, Brian Ichter, Ayzaan Wahid, Jonathan Tompson, Quan Vuong, Tianhe Yu, et~al.
\newblock Palm-e: An embodied multimodal language model.
\newblock \emph{arXiv preprint arXiv:2303.03378}, 2023.

\bibitem[Fan et~al.(2018)Fan, Lewis, and Dauphin]{fan2018topk}
Angela Fan, Mike Lewis, and Yann Dauphin.
\newblock Hierarchical neural story generation.
\newblock \emph{arXiv preprint arXiv:1805.04833}, 2018.

\bibitem[Fu et~al.(2023)Fu, Li, Wen, Dou, Cai, Shi, and Qiao]{fu2023drive}
Daocheng Fu, Xin Li, Licheng Wen, Min Dou, Pinlong Cai, Botian Shi, and Yu Qiao.
\newblock Drive like a human: Rethinking autonomous driving with large language models.
\newblock \emph{arXiv preprint arXiv:2307.07162}, 2023.

\bibitem[Guo et~al.(2023)Guo, Zhang, Zhu, Tang, Ma, Han, Chen, Gao, Li, Li, et~al.]{guo2023pointbind+pointllm}
Ziyu Guo, Renrui Zhang, Xiangyang Zhu, Yiwen Tang, Xianzheng Ma, Jiaming Han, Kexin Chen, Peng Gao, Xianzhi Li, Hongsheng Li, et~al.
\newblock Point-bind \& point-llm: Aligning point cloud with multi-modality for 3d understanding, generation, and instruction following.
\newblock \emph{arXiv preprint arXiv:2309.00615}, 2023.

\bibitem[Holtzman et~al.(2019)Holtzman, Buys, Du, Forbes, and Choi]{holtzman2019topp}
Ari Holtzman, Jan Buys, Li Du, Maxwell Forbes, and Yejin Choi.
\newblock The curious case of neural text degeneration.
\newblock \emph{arXiv preprint arXiv:1904.09751}, 2019.

\bibitem[Hong et~al.(2023{\natexlab{a}})Hong, Lin, Du, Chen, Tenenbaum, and Gan]{hong20233d}
Yining Hong, Chunru Lin, Yilun Du, Zhenfang Chen, Joshua~B Tenenbaum, and Chuang Gan.
\newblock 3d concept learning and reasoning from multi-view images.
\newblock In \emph{Proceedings of the IEEE/CVF Conference on Computer Vision and Pattern Recognition}, pages 9202--9212, 2023{\natexlab{a}}.

\bibitem[Hong et~al.(2023{\natexlab{b}})Hong, Zhen, Chen, Zheng, Du, Chen, and Gan]{hong20233d-llm}
Yining Hong, Haoyu Zhen, Peihao Chen, Shuhong Zheng, Yilun Du, Zhenfang Chen, and Chuang Gan.
\newblock 3d-llm: Injecting the 3d world into large language models.
\newblock \emph{arXiv preprint arXiv:2307.12981}, 2023{\natexlab{b}}.

\bibitem[Iyer et~al.(2022)Iyer, Lin, Pasunuru, Mihaylov, Simig, Yu, Shuster, Wang, Liu, Koura, et~al.]{iyer2022opt-iml}
Srinivasan Iyer, Xi~Victoria Lin, Ramakanth Pasunuru, Todor Mihaylov, Daniel Simig, Ping Yu, Kurt Shuster, Tianlu Wang, Qing Liu, Punit~Singh Koura, et~al.
\newblock Opt-iml: Scaling language model instruction meta learning through the lens of generalization.
\newblock \emph{arXiv preprint arXiv:2212.12017}, 2022.

\bibitem[Jiang et~al.(2023)Jiang, Chen, Liu, Yu, Yu, and Chen]{jiang2023motiongpt}
Biao Jiang, Xin Chen, Wen Liu, Jingyi Yu, Gang Yu, and Tao Chen.
\newblock Motiongpt: Human motion as a foreign language.
\newblock \emph{arXiv preprint arXiv:2306.14795}, 2023.

\bibitem[Jiao et~al.(2022)Jiao, Chen, Jie, Chen, Ma, and Jiang]{jiao2022more}
Yang Jiao, Shaoxiang Chen, Zequn Jie, Jingjing Chen, Lin Ma, and Yu-Gang Jiang.
\newblock More: Multi-order relation mining for dense captioning in 3d scenes.
\newblock \emph{arXiv preprint arXiv:2203.05203}, 2022.

\bibitem[Jin et~al.(2023)Jin, Hayat, Yang, Guo, and Lei]{jin2023-3D-VLP}
Zhao Jin, Munawar Hayat, Yuwei Yang, Yulan Guo, and Yinjie Lei.
\newblock Context-aware alignment and mutual masking for 3d-language pre-training.
\newblock In \emph{Proceedings of the IEEE/CVF Conference on Computer Vision and Pattern Recognition}, pages 10984--10994, 2023.

\bibitem[Kirillov et~al.(2023)Kirillov, Mintun, Ravi, Mao, Rolland, Gustafson, Xiao, Whitehead, Berg, Lo, Doll{\'a}r, and Girshick]{kirillov2023segany}
Alexander Kirillov, Eric Mintun, Nikhila Ravi, Hanzi Mao, Chloe Rolland, Laura Gustafson, Tete Xiao, Spencer Whitehead, Alexander~C. Berg, Wan-Yen Lo, Piotr Doll{\'a}r, and Ross Girshick.
\newblock Segment anything.
\newblock \emph{arXiv:2304.02643}, 2023.

\bibitem[Li et~al.(2023)Li, Li, Savarese, and Hoi]{li2023blip2}
Junnan Li, Dongxu Li, Silvio Savarese, and Steven Hoi.
\newblock Blip-2: Bootstrapping language-image pre-training with frozen image encoders and large language models.
\newblock \emph{arXiv preprint arXiv:2301.12597}, 2023.

\bibitem[Lin(2004)]{lin2004rouge}
Chin-Yew Lin.
\newblock Rouge: A package for automatic evaluation of summaries.
\newblock In \emph{Text summarization branches out}, pages 74--81, 2004.

\bibitem[Liu et~al.(2023)Liu, Li, Wu, and Lee]{liu2023llava}
Haotian Liu, Chunyuan Li, Qingyang Wu, and Yong~Jae Lee.
\newblock Visual instruction tuning.
\newblock \emph{arXiv preprint arXiv:2304.08485}, 2023.

\bibitem[Loshchilov and Hutter(2017)]{loshchilov2017AdamW}
Ilya Loshchilov and Frank Hutter.
\newblock Decoupled weight decay regularization.
\newblock \emph{arXiv preprint arXiv:1711.05101}, 2017.

\bibitem[Luo et~al.(2023)Luo, Rockwell, Lee, and Johnson]{luo2023cap3d}
Tiange Luo, Chris Rockwell, Honglak Lee, and Justin Johnson.
\newblock Scalable 3d captioning with pretrained models.
\newblock \emph{arXiv preprint arXiv:2306.07279}, 2023.

\bibitem[Ma et~al.(2022)Ma, Yong, Zheng, Li, Liang, Zhu, and Huang]{ma2022sqa3d}
Xiaojian Ma, Silong Yong, Zilong Zheng, Qing Li, Yitao Liang, Song-Chun Zhu, and Siyuan Huang.
\newblock Sqa3d: Situated question answering in 3d scenes.
\newblock \emph{arXiv preprint arXiv:2210.07474}, 2022.

\bibitem[Mao et~al.(2023)Mao, Yang, Chen, Yi, and Liu]{mao2023REMAN}
Aihua Mao, Zhi Yang, Wanxin Chen, Ran Yi, and Yong-jin Liu.
\newblock Complete 3d relationships extraction modality alignment network for 3d dense captioning.
\newblock \emph{IEEE Transactions on Visualization and Computer Graphics}, 2023.

\bibitem[Misra et~al.(2021)Misra, Girdhar, and Joulin]{misra2021-3detr}
Ishan Misra, Rohit Girdhar, and Armand Joulin.
\newblock An end-to-end transformer model for 3d object detection.
\newblock In \emph{Proceedings of the IEEE/CVF International Conference on Computer Vision}, pages 2906--2917, 2021.

\bibitem[Mokady et~al.(2021)Mokady, Hertz, and Bermano]{mokady2021clipcap}
Ron Mokady, Amir Hertz, and Amit~H Bermano.
\newblock Clipcap: Clip prefix for image captioning.
\newblock \emph{arXiv preprint arXiv:2111.09734}, 2021.

\bibitem[OpenAI(2023)]{openai2023gpt4}
OpenAI.
\newblock Gpt-4 technical report, 2023.

\bibitem[Papineni et~al.(2002)Papineni, Roukos, Ward, and Zhu]{papineni2002bleu}
Kishore Papineni, Salim Roukos, Todd Ward, and Wei-Jing Zhu.
\newblock Bleu: a method for automatic evaluation of machine translation.
\newblock In \emph{Proceedings of the 40th annual meeting of the Association for Computational Linguistics}, pages 311--318, 2002.

\bibitem[Parelli et~al.(2023)Parelli, Delitzas, Hars, Vlassis, Anagnostidis, Bachmann, and Hofmann]{parelli2023clip-guided}
Maria Parelli, Alexandros Delitzas, Nikolas Hars, Georgios Vlassis, Sotirios Anagnostidis, Gregor Bachmann, and Thomas Hofmann.
\newblock Clip-guided vision-language pre-training for question answering in 3d scenes.
\newblock In \emph{Proceedings of the IEEE/CVF Conference on Computer Vision and Pattern Recognition}, pages 5606--5611, 2023.

\bibitem[Qi et~al.(2017)Qi, Yi, Su, and Guibas]{qi2017pointnet++}
Charles~Ruizhongtai Qi, Li Yi, Hao Su, and Leonidas~J Guibas.
\newblock Pointnet++: Deep hierarchical feature learning on point sets in a metric space.
\newblock \emph{Advances in neural information processing systems}, 30, 2017.

\bibitem[Qi et~al.(2019)Qi, Litany, He, and Guibas]{qi2019votenet}
Charles~R Qi, Or Litany, Kaiming He, and Leonidas~J Guibas.
\newblock Deep hough voting for 3d object detection in point clouds.
\newblock In \emph{proceedings of the IEEE/CVF International Conference on Computer Vision}, pages 9277--9286, 2019.

\bibitem[Rozenberszki et~al.(2022)Rozenberszki, Litany, and Dai]{rozenberszki2022language}
David Rozenberszki, Or Litany, and Angela Dai.
\newblock Language-grounded indoor 3d semantic segmentation in the wild.
\newblock In \emph{Proceedings of the European Conference on Computer Vision ({ECCV})}, 2022.

\bibitem[Song et~al.(2023)Song, Wu, Washington, Sadler, Chao, and Su]{song2023llm}
Chan~Hee Song, Jiaman Wu, Clayton Washington, Brian~M Sadler, Wei-Lun Chao, and Yu Su.
\newblock Llm-planner: Few-shot grounded planning for embodied agents with large language models.
\newblock In \emph{Proceedings of the IEEE/CVF International Conference on Computer Vision}, pages 2998--3009, 2023.

\bibitem[Tancik et~al.(2020)Tancik, Srinivasan, Mildenhall, Fridovich-Keil, Raghavan, Singhal, Ramamoorthi, Barron, and Ng]{tancik2020fourier}
Matthew Tancik, Pratul Srinivasan, Ben Mildenhall, Sara Fridovich-Keil, Nithin Raghavan, Utkarsh Singhal, Ravi Ramamoorthi, Jonathan Barron, and Ren Ng.
\newblock Fourier features let networks learn high frequency functions in low dimensional domains.
\newblock \emph{Advances in Neural Information Processing Systems}, 33:\penalty0 7537--7547, 2020.

\bibitem[Touvron et~al.(2023)Touvron, Martin, Stone, Albert, Almahairi, Babaei, Bashlykov, Batra, Bhargava, Bhosale, et~al.]{touvron2023llama}
Hugo Touvron, Louis Martin, Kevin Stone, Peter Albert, Amjad Almahairi, Yasmine Babaei, Nikolay Bashlykov, Soumya Batra, Prajjwal Bhargava, Shruti Bhosale, et~al.
\newblock Llama 2: Open foundation and fine-tuned chat models.
\newblock \emph{arXiv preprint arXiv:2307.09288}, 2023.

\bibitem[Vedantam et~al.(2015)Vedantam, Lawrence~Zitnick, and Parikh]{vedantam2015cider}
Ramakrishna Vedantam, C Lawrence~Zitnick, and Devi Parikh.
\newblock Cider: Consensus-based image description evaluation.
\newblock In \emph{Proceedings of the IEEE conference on computer vision and pattern recognition}, pages 4566--4575, 2015.

\bibitem[Wang et~al.(2022)Wang, Zhang, Yu, and Cai]{wang2022spacap3d}
Heng Wang, Chaoyi Zhang, Jianhui Yu, and Weidong Cai.
\newblock Spatiality-guided transformer for 3d dense captioning on point clouds.
\newblock \emph{arXiv preprint arXiv:2204.10688}, 2022.

\bibitem[Wu et~al.(2023)Wu, Cheng, Zhang, Cheng, and Zhang]{wu2023eda}
Yanmin Wu, Xinhua Cheng, Renrui Zhang, Zesen Cheng, and Jian Zhang.
\newblock Eda: Explicit text-decoupling and dense alignment for 3d visual grounding.
\newblock In \emph{Proceedings of the IEEE/CVF Conference on Computer Vision and Pattern Recognition}, pages 19231--19242, 2023.

\bibitem[Xu et~al.(2023{\natexlab{a}})Xu, Zhu, and Clifton]{xu2023survey-multimodal}
Peng Xu, Xiatian Zhu, and David~A Clifton.
\newblock Multimodal learning with transformers: A survey.
\newblock \emph{IEEE Transactions on Pattern Analysis and Machine Intelligence}, 2023{\natexlab{a}}.

\bibitem[Xu et~al.(2023{\natexlab{b}})Xu, Wang, Wang, Chen, Pang, and Lin]{xu2023pointllm}
Runsen Xu, Xiaolong Wang, Tai Wang, Yilun Chen, Jiangmiao Pang, and Dahua Lin.
\newblock Pointllm: Empowering large language models to understand point clouds.
\newblock \emph{arXiv preprint arXiv:2308.16911}, 2023{\natexlab{b}}.

\bibitem[Ye et~al.(2023)Ye, Xu, Xu, Ye, Yan, Zhou, Wang, Hu, Shi, Shi, et~al.]{ye2023mplug-owl}
Qinghao Ye, Haiyang Xu, Guohai Xu, Jiabo Ye, Ming Yan, Yiyang Zhou, Junyang Wang, Anwen Hu, Pengcheng Shi, Yaya Shi, et~al.
\newblock mplug-owl: Modularization empowers large language models with multimodality.
\newblock \emph{arXiv preprint arXiv:2304.14178}, 2023.

\bibitem[Ye et~al.(2022)Ye, Chen, Han, and Liao]{ye20223d}
Shuquan Ye, Dongdong Chen, Songfang Han, and Jing Liao.
\newblock 3d question answering.
\newblock \emph{IEEE Transactions on Visualization and Computer Graphics}, 2022.

\bibitem[Yin et~al.(2023)Yin, Fu, Zhao, Li, Sun, Xu, and Chen]{yin2023survey-mllm}
Shukang Yin, Chaoyou Fu, Sirui Zhao, Ke Li, Xing Sun, Tong Xu, and Enhong Chen.
\newblock A survey on multimodal large language models.
\newblock \emph{arXiv preprint arXiv:2306.13549}, 2023.

\bibitem[Zhang et~al.(2022)Zhang, Roller, Goyal, Artetxe, Chen, Chen, Dewan, Diab, Li, Lin, et~al.]{zhang2022opt}
Susan Zhang, Stephen Roller, Naman Goyal, Mikel Artetxe, Moya Chen, Shuohui Chen, Christopher Dewan, Mona Diab, Xian Li, Xi~Victoria Lin, et~al.
\newblock Opt: Open pre-trained transformer language models.
\newblock \emph{arXiv preprint arXiv:2205.01068}, 2022.

\bibitem[Zhang et~al.(2023)Zhang, Sun, Chen, Xiao, Shao, Zhang, Chen, and Luo]{zhang2023gpt4roi}
Shilong Zhang, Peize Sun, Shoufa Chen, Min Xiao, Wenqi Shao, Wenwei Zhang, Kai Chen, and Ping Luo.
\newblock Gpt4roi: Instruction tuning large language model on region-of-interest.
\newblock \emph{arXiv preprint arXiv:2307.03601}, 2023.

\bibitem[Zhao et~al.(2021)Zhao, Cai, Sheng, and Xu]{zhao20213dvg}
Lichen Zhao, Daigang Cai, Lu Sheng, and Dong Xu.
\newblock 3dvg-transformer: Relation modeling for visual grounding on point clouds.
\newblock In \emph{Proceedings of the IEEE/CVF International Conference on Computer Vision}, pages 2928--2937, 2021.

\bibitem[Zhao et~al.(2022)Zhao, Cai, Zhang, Sheng, Xu, Zheng, Zhao, Wang, and Fan]{zhao2022-fe-3dgqa}
Lichen Zhao, Daigang Cai, Jing Zhang, Lu Sheng, Dong Xu, Rui Zheng, Yinjie Zhao, Lipeng Wang, and Xibo Fan.
\newblock Towards explainable 3d grounded visual question answering: A new benchmark and strong baseline.
\newblock \emph{IEEE Transactions on Circuits and Systems for Video Technology}, 2022.

\bibitem[Zhong et~al.(2022)Zhong, Xu, Luo, and Ma]{zhong2022contextual3DdenseCap}
Yufeng Zhong, Long Xu, Jiebo Luo, and Lin Ma.
\newblock Contextual modeling for 3d dense captioning on point clouds.
\newblock \emph{arXiv preprint arXiv:2210.03925}, 2022.

\bibitem[Zhou et~al.(2023)Zhou, Yu, Zhang, Wu, Wang, and Wang]{zhou2023regionblip}
Qiang Zhou, Chaohui Yu, Shaofeng Zhang, Sitong Wu, Zhibing Wang, and Fan Wang.
\newblock Regionblip: A unified multi-modal pre-training framework for holistic and regional comprehension, 2023.

\bibitem[Zhu et~al.(2023{\natexlab{a}})Zhu, Chen, Shen, Li, and Elhoseiny]{zhu2023minigpt4}
Deyao Zhu, Jun Chen, Xiaoqian Shen, Xiang Li, and Mohamed Elhoseiny.
\newblock Minigpt-4: Enhancing vision-language understanding with advanced large language models.
\newblock \emph{arXiv preprint arXiv:2304.10592}, 2023{\natexlab{a}}.

\bibitem[Zhu et~al.(2023{\natexlab{b}})Zhu, Ma, Chen, Deng, Huang, and Li]{zhu20233d-vista}
Ziyu Zhu, Xiaojian Ma, Yixin Chen, Zhidong Deng, Siyuan Huang, and Qing Li.
\newblock 3d-vista: Pre-trained transformer for 3d vision and text alignment.
\newblock In \emph{Proceedings of the IEEE/CVF International Conference on Computer Vision}, pages 2911--2921, 2023{\natexlab{b}}.

\end{thebibliography}
}

\clearpage
\onecolumn
\section*{
    {\hfill \LARGE Supplementary Material \hfill }
}
\vspace{100pt}
\renewcommand\thesection{\Alph{section}}
\setcounter{section}{0}
\noindent The supplementary material consists of more visualization results (\cref{appendix-sec:more-visualizations}), quantitative evaluations on scene descriptions, embodied dialogue, and embodied planning (\cref{appendix-sec:more-evaluations}), additional studies on 3D coordinates generated by \modelname{} (\cref{appendix-sec:additional studies-llm-boxes}), data statistics as well as training samples (\cref{appendix-sec:data}), additional implementation details (\cref{appendix-sec:implementation}).

\myparagraph{Codes and Pre-trained weights} will be fully released.
We provide example code files along with this file, including the training and evaluation process for different tasks.

\myparagraph{Video.} We also provide video clips for more illustrative demonstrations.

\section{More Visualizations}
\label{appendix-sec:more-visualizations}

\begin{figure*}[htbp]
	\centering
	\includegraphics[width=\linewidth]{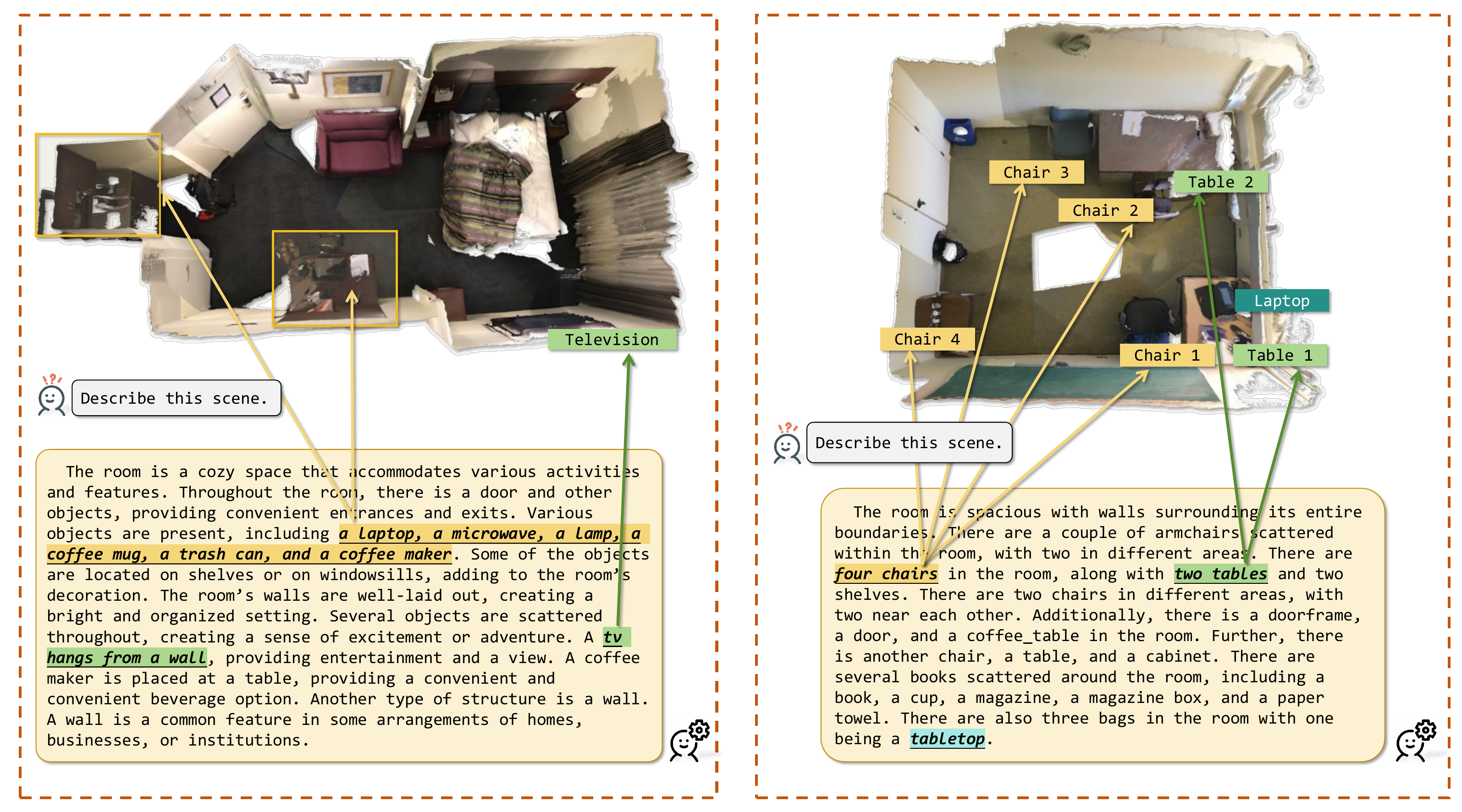}
	\caption{
    	\textbf{Qualitative Results on Scene Descriptions.}
        We highlight some of the phrases in the generated scene descriptions mentioning the instances in the 3D environment.
    }
	\label{appendix-fig:scene description}
\end{figure*}

\begin{figure*}[htbp]
	\centering
	\includegraphics[width=\linewidth]{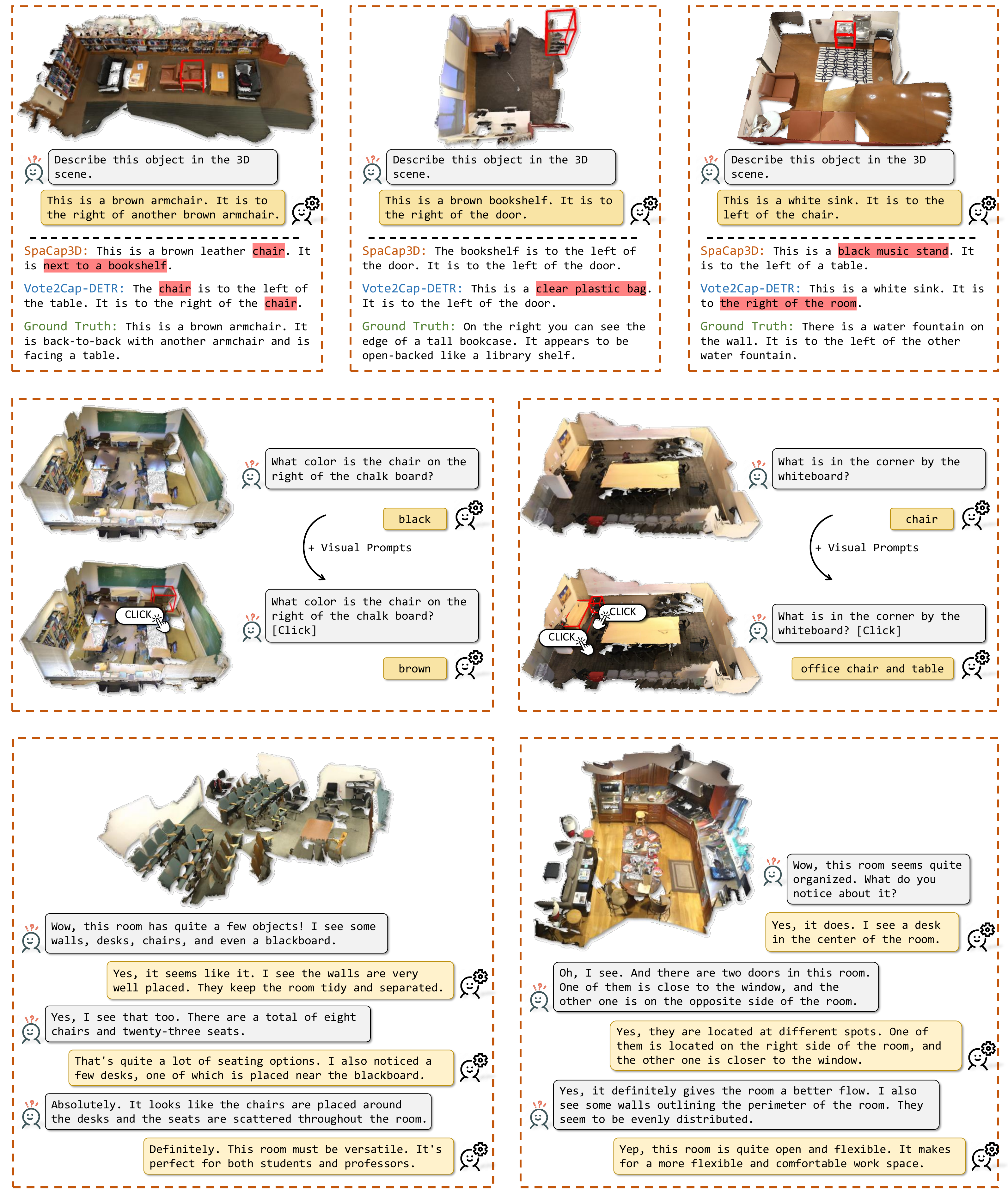}
	\caption{
    	\textbf{More Qualitative Results on 3D Dense Captioning (upper), 3D Question Answering (middle), and Embodied Dialogue (lower).}
        \colorbox[RGB]{255, 124, 128}{Red} highlights the wrong answer. 
    }
	\label{appendix-fig:3D dc + qa + ed}
\end{figure*}

\begin{figure*}[htbp]
	\centering
	\includegraphics[width=0.95\linewidth]{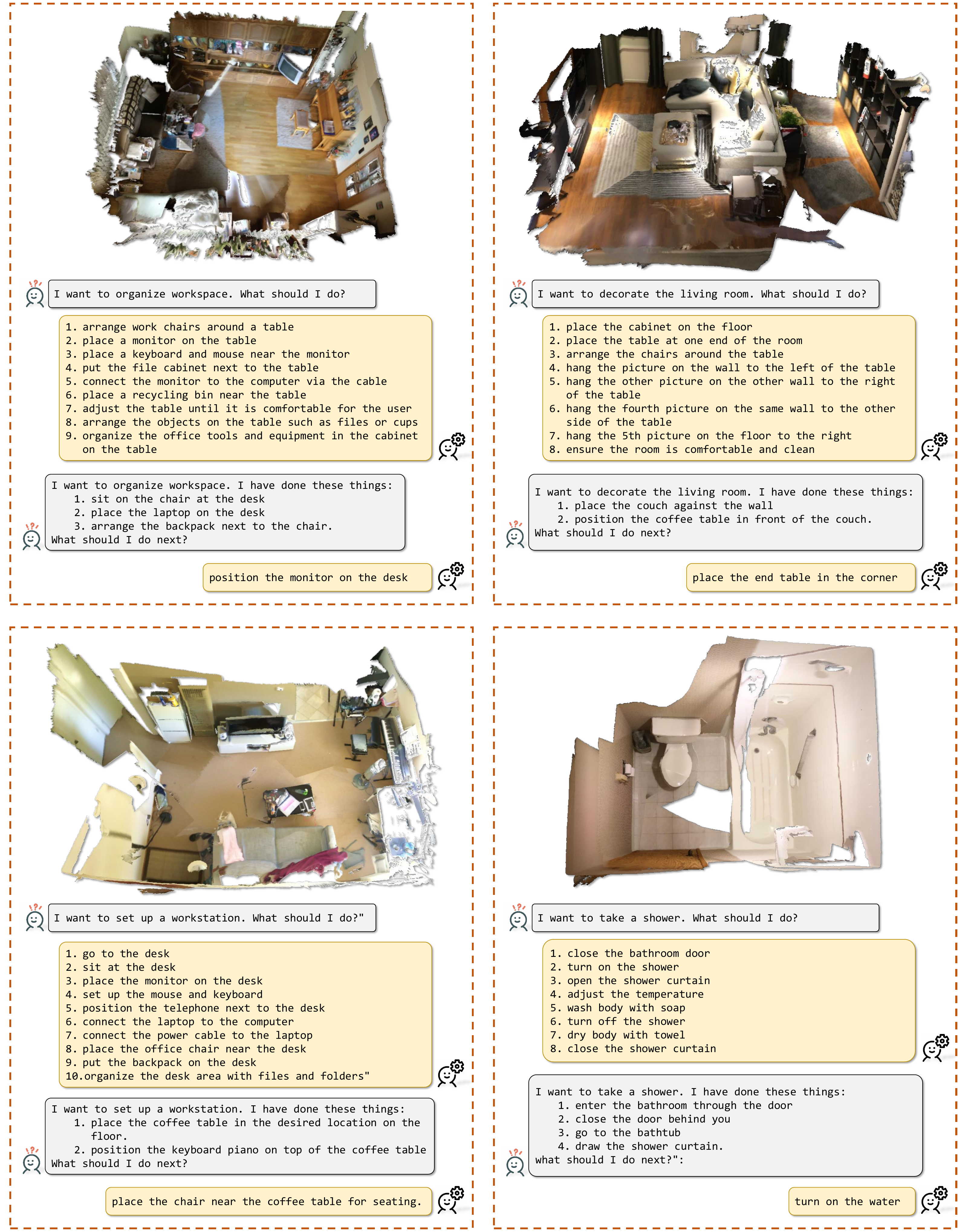}
	\caption{
    	\textbf{Qualitative Results on Embodied Planning.}
        The planning results generated by \modelname{} are consistent with common sense knowledge.
        Additionally, \modelname{} could not only make plans directly, but also generate feedbacks based on the things have been done.
    }
	\label{appendix-fig:embodied planning}
\end{figure*}

We present additional visualization results on Scene Descriptions (\cref{appendix-fig:scene description}), 3D Dense Captioning, 3D Question Answering, Embodied Dialogue (\cref{appendix-fig:3D dc + qa + ed}), and Embodied Planning (\cref{appendix-fig:embodied planning}).


\clearpage

\section{More Evaluations}
\label{appendix-sec:more-evaluations}

Since 3D-LLM \cite{hong20233d-llm} has not yet released the validation set they used for evaluation, we make our own split to quantify our method's performance on: 1) scene descriptions, 2) embodied dialogue, and 3) embodied planning.

\myparagraph{Data Splits.} In practice, we set the scenes with ids less than 600 as the training set, and the rest as the validation set.
The data statistics of our split are listed in \cref{appendix-tab:3d-llm data statistics}.
It is worth mentioning that:
\begin{itemize} 
\setlength\itemsep{0em}
    \item The multi-turn embodied dialogues are further decomposed into 8,490 and 1,222 lines of data samples following the procedure in \cref{appendix-fig:dialogue-decomposition} for training and validation.
    \item The embodied planning data is also decomposed into 16,972 and 2,282 lines of data samples for training and validation following the procedure in \cref{appendix-fig:planning-decomposition}.
\end{itemize}
As mentioned in Sec. 4.1 in the main paper, we distinguishes the source of texts with the ``\textcolor[RGB]{215,48,39}{\#\#\# human:}'' and ``\textcolor[RGB]{69,117,180}{\#\#\# assistant:}'' identifier.

\myparagraph{Metrics.} Similar to the main paper, we evaluate the generated natural language responses under the CiDEr \cite{vedantam2015cider}, BLEU \cite{papineni2002bleu}, METEOR \cite{banerjee2005meteor}, and Rouge-L \cite{lin2004rouge} metrics in \cref{appendix-tab:3d-llm}.

\begin{figure*}[htbp]
	\centering
	\includegraphics[width=\linewidth]{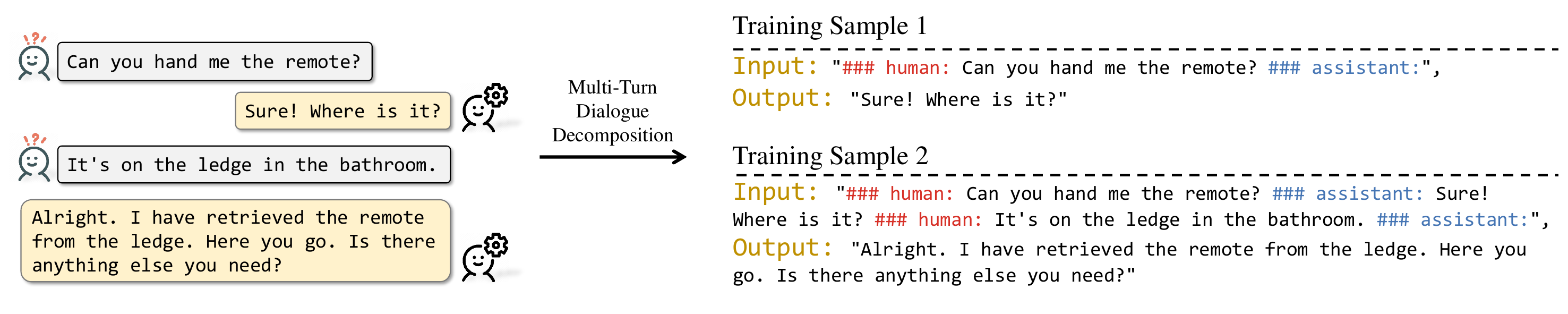}
    \vspace{-5mm}
	\caption{
    	\textbf{The Multi-Turn Dialogue Decomposition Pipeline.}
        For one sample of a $n$-turn dialogue data ($n \ge 1$), we decompose it into $n$ lines of training samples, and distinguish the source of data with the ``\textcolor[RGB]{215,48,39}{\#\#\# human:}'' and ``\textcolor[RGB]{69,117,180}{\#\#\# assistant:}'' identifier.
    }
	\label{appendix-fig:dialogue-decomposition}
\end{figure*}
\begin{figure*}[htbp]
	\centering
	\includegraphics[width=\linewidth]{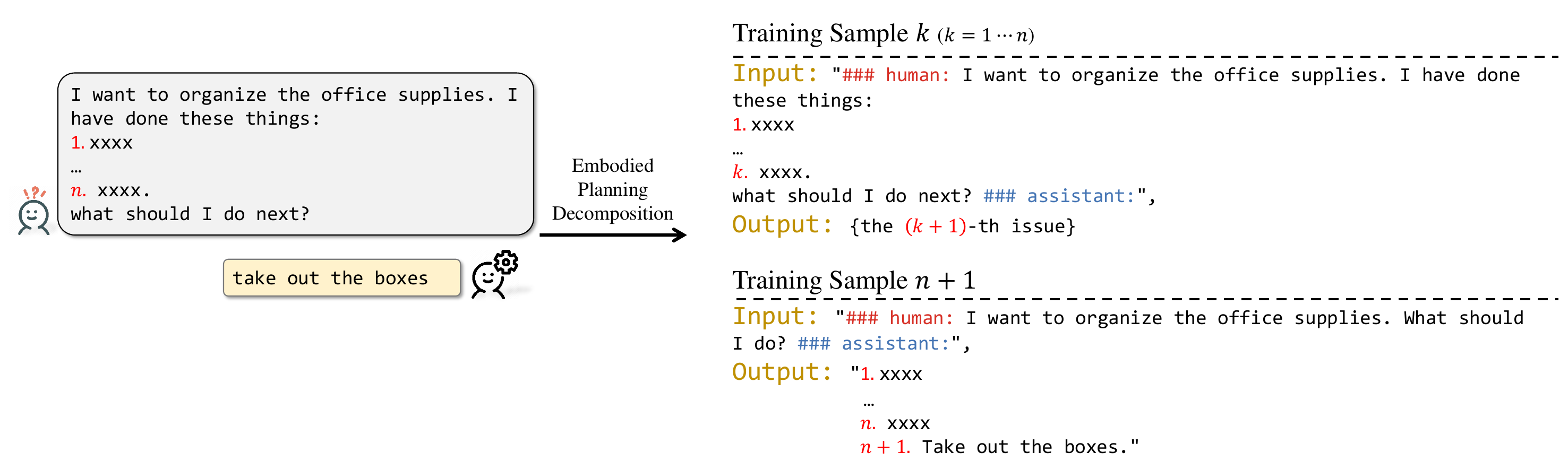}
    \vspace{-5mm}
	\caption{
    	\textbf{The Embodied Planning Decomposition Pipeline.}
        For one sample with given $n$-tasks ($n \ge 1$), we decompose it into $(n+1)$ lines of training and validation samples.
    }
	\label{appendix-fig:planning-decomposition}
\end{figure*}

\begin{table*}[htbp]
\caption{
        \textbf{Details about the Training-Validation Split of the ScanNet Part of 3D-LLM \cite{hong20233d-llm}.}
        We list the number of natural language annotations (``\# annotations'') and number of scenes (``\# scenes'') in each split for each task.
    }
    \label{appendix-tab:3d-llm data statistics}
    \centering
    \resizebox{0.75\linewidth}{!}{
    \begin{tabular}{ccccccccc}
    \toprule
    \multirow{2}{*}{Split} & \multicolumn{2}{c}{Scene Description} &  & \multicolumn{2}{c}{Embodied Planning} &  & \multicolumn{2}{c}{Embodied Dialogue} \\ \cline{2-3} \cline{5-6} \cline{8-9} 
                           & \# annotations       & \# scenes      &  & \# annotations       & \# scenes      &  & \# annotations       & \# scenes      \\ \hline
    Train                  & 912                  & 456            &  & 1,636                & 449            &  & 2,592                & 456            \\
    Validation             & 121                  & 61             &  & 228                  & 61             &  & 363                  & 61             \\ \hline
    Total                  & 1,033                & 517            &  & 1,864                & 510            &  & 2,955                & 517            \\ 
    \bottomrule
    \end{tabular}
    }
\end{table*}

\begin{table*}[htbp]
    \caption{
        \textbf{Quantitative Comparisons on Scene Description, Embodied Dialogue and Embodied Planning.}
        We evaluate our method with different sampling strategies on the ScanNet part of 3D-LLM \cite{hong20233d-llm}.
        We manually set the scenes with ids less than 600 as the training set, and the rest as the validation set.
        To prevent repetition, we adopt the $n$-gram repetition penalty so that no $n$-gram appears twice ($n=4$).
    }
    \label{appendix-tab:3d-llm}
    \centering
    \resizebox{0.8\linewidth}{!}{
    \begin{tabular}{ccccccccc}
    \toprule
    Task                               & Method    & BLEU-1$\uparrow$ & BLEU-2$\uparrow$ & BLEU-3$\uparrow$ & BLEU-4$\uparrow$ & CiDEr$\uparrow$ & METEOR$\uparrow$ & Rouge-L$\uparrow$ \\ \hline
    %
    %
    %
    %
    \multirow{9}{*}{\begin{tabular}[c]{@{}c@{}}Scene\\Description\end{tabular}} 
    & \multicolumn{1}{l}{\textbf{\textit{Zero-Shot:}}} \\
    & OPT-1.3B \cite{zhang2022opt} & 15.79            &  6.10            &  2.07            &  0.84              & 0.00           &  8.40            & 11.70             \\
    & OPT-2.7B \cite{zhang2022opt} & 19.97            &  7.59            &  2.14            &  0.00             &  0.11          & 6.60             & 12.32             \\
    & OPT-6.7B \cite{zhang2022opt} & 24.40            &  9.79            &  3.62            &  1.13             &  0.06          &  8.99            & 16.96             \\
    & LLAMA-7B \cite{touvron2023llama} & 19.26            &  7.69            &  2.79            &  0.92             &  0.20          &  7.00            & 12.31             \\
    \cline{2-2}
    & \multicolumn{1}{l}{\textbf{\textit{Ours:}}} \\
    & top-$k$ \& top-$p$ & 43.02            & 26.70            & 15.97            &   8.97            & 0.96           & 14.65            & 24.84             \\
    & greedy decoding    & 29.15            & 20.51            & 13.99            &  9.38            & 1.44           & 12.83            & 24.62             \\ 
    & beam search        & 29.94            & 21.56            & 14.93            & 10.02            & 1.32           & 12.31            & 27.08             \\ 
    \hline
    %
    %
    %
    %
    \multirow{9}{*}{\begin{tabular}[c]{@{}c@{}}Embodied\\Dialogue\end{tabular}}
    & \multicolumn{1}{l}{\textbf{\textit{Zero-Shot:}}} \\
    & OPT-1.3B \cite{zhang2022opt} &  2.44            &  1.05            &  0.46            &  0.23             &  0.31          &  5.62            &  4.83             \\
    & OPT-2.7B \cite{zhang2022opt} &  3.88            &  1.56            &  0.73            &  0.39             &  0.38          &  7.38            &  6.28             \\
    & OPT-6.7B \cite{zhang2022opt} & 3.59            &  1.65            &  0.81            &  0.43             &  0.25          &  6.88            &  6.16             \\
    & LLAMA-7B \cite{touvron2023llama} &  4.08            &  1.80            &  0.90            &  0.50             &  0.27          &  7.81            &  6.68             \\
    \cline{2-2}
    & \multicolumn{1}{l}{\textbf{\textit{Ours:}}} \\
    & top-$k$ \& top-$p$ & 41.38            & 32.59            & 27.47            & 23.95            & 190.01           & 23.50            & 40.61             \\
    & greedy decoding    & 48.08            & 39.59            & 34.37            & 30.70            & 251.78           & 27.01            & 47.57             \\ 
    & beam search        & 48.14            & 39.83            & 34.83            & 31.32            & 260.07           & 27.21            & 47.69             \\ 
    \hline
    %
    %
    %
    %
    \multirow{9}{*}{\begin{tabular}[c]{@{}c@{}}Embodied\\Planning\end{tabular}}
    & \multicolumn{1}{l}{\textbf{\textit{Zero-Shot:}}} \\
    & OPT-1.3B \cite{zhang2022opt} &  1.26            &  0.59            &  0.26            &  0.13             &  0.16          &  0.24            &  3.56             \\
    & OPT-2.7B \cite{zhang2022opt} & 2.02             &  0.99            &  0.49            &  0.26             &  0.10          &  3.59            &  4.35             \\
    & OPT-6.7B \cite{zhang2022opt} &  2.03            &  1.06            &  0.53            &  0.28             &  0.00          &  3.65            &  3.94             \\
    & LLAMA-7B \cite{touvron2023llama} &  2.24            &  1.13            &  0.55            &  0.29             &  0.04          &  3.53            &  4.71             \\
    \cline{2-2}
    & \multicolumn{1}{l}{\textbf{\textit{Ours:}}} \\
    & top-$k$ \& top-$p$ & 40.72            & 27.18            & 18.64            & 12.95            & 128.80           & 17.05            & 39.25             \\
    & greedy decoding    & 38.81            & 27.58            & 20.10            & 14.66            & 186.13           & 19.60            & 45.34             \\ 
    & beam search        & 45.07            & 33.04            & 24.96            & 19.15            & 196.78           & 19.87            & 45.58             \\ \bottomrule
    \end{tabular}
    }
\end{table*}

\myparagraph{Quantitative Comparisons.}
As shown in \cref{appendix-tab:3d-llm}, we perform zero-shot evaluation on the frozen LLMs \cite{zhang2022opt,touvron2023llama} with the top-$k$ and top-$p$ sampling strategy as the baseline, and evaluate our method with different sampling strategies.
Among all the listed results, we adopt the $n$-gram penalty that restricts no $n$-gram appears twice with $n=4$.
We also set $k=50$ and $p=0.95$ for top-$k$ and top-$p$ sampling, and a beam size $=4$ for beam search.
From the table, one could see that our method could generate high-quality responses, and out-performs language-only LLMs.

\clearpage

\section{Additional Studies: Generating 3D Bounding Boxes with LLM}
\label{appendix-sec:additional studies-llm-boxes}

\myparagraph{Quantitative Evaluation on 3D Open-Vocabulary Detection with LLM-Generated Boxes.} 
We treat 3D Open-Vocabulary Detection as the 3D Dense Captioning problem, and simulate the click prompts using the spatial location of vote queries \cite{chen2023vote2cap-detr}.
The textual instruction is set as ``\underline{\textit{\textcolor[RGB]{215,48,39}{\#\#\# human:} what is this object? \textcolor[RGB]{69,117,180}{\#\#\# assistant:}}}''.
To obtain localization results, we reconstruct 3D bounding boxes from the ``\obj{$c_x$, $c_y$, $c_z$, $w$, $h$, $l$}'' in the generated texts, and also decode the object category prediction from the texts.
In \cref{appendix-tab:scanrefer llm boxes}, we present performances under the ScanNet vocabulary \cite{dai2017scannet}, and the ScanNet200 vocabulary \cite{rozenberszki2022language}.
Per-class mAP results on ScanNet could be found in \cref{appendix-tab:scannet AP per class}.
Results show that \modelname{} could produce high-quality bounding boxes comparable to 3D detectors under the threshold IoU=0.5.

\begin{table*}[htbp]
    \caption{
        \textbf{Quantitative Comparisons on Open-Vocabulary Detection with \modelname{}.}
        We treat 3D Detection as the 3D Dense Captioning problem.
        We simulate the click prompt with the vote queries \cite{chen2023vote2cap-detr}, and decode the ``\obj{$c_x$, $c_y$, $c_z$, $w$, $h$, $l$}'' and category names from the generated texts.
        Results show that our method is able to produce comparable to the 3D detector specialist when IoU=0.5.
    }
    \label{appendix-tab:open vocabulary detection}
    \centering
    \resizebox{0.7\linewidth}{!}{
    \begin{tabular}{c|ccccccccccc}
    \toprule
    \multirow{3}{*}{Method} & \multicolumn{5}{c}{ScanNet Vocabulary}                         &  & \multicolumn{5}{c}{ScanNet200 Vocabulary}\\ \cline{2-6} \cline{8-12}
                        & \multicolumn{2}{c}{IoU=0.25} &  & \multicolumn{2}{c}{IoU=0.5}  &  & \multicolumn{2}{c}{IoU=0.25} &  & \multicolumn{2}{c}{IoU=0.5}  \\ \cline{2-3} \cline{5-6} \cline{8-9} \cline{11-12}
                        & mAP$\uparrow$ & AR$\uparrow$ &  & mAP$\uparrow$ & AR$\uparrow$ &  & mAP$\uparrow$ & AR$\uparrow$ &  & mAP$\uparrow$ & AR$\uparrow$  \\ \hline
    VoteNet \cite{qi2019votenet} & 57.17         & 81.18        &  & 31.50         & 50.08        &  & -             & -            &  & -             & -            \\
    Ours                    & 48.94         & 65.15        &  & 32.48         & 49.46        &  & 7.40         & 12.10        &  & 5.20         & 9.04           \\
    \bottomrule
    \end{tabular}
}
\end{table*}

\begin{table*}[htbp]
    \centering
    \caption{\textbf{Per-class AP under IoU threshold of 0.25 and 0.5 on ScanNet validation set.}}
    \label{appendix-tab:scannet AP per class}
    \resizebox{\linewidth}{!}{
    \begin{tabular}{c|cccccccccccccccccc}
    \toprule
    IoU           & bathtub & bed   & bookshelf & cabinet  & chair & counter  & curtain & desk & door & others & picture  & refrigerator & shower curtain & sink & sofa & table  & toilet & window \\ \hline
    0.25          & 83.33   & 62.16 & 10.63 & 43.13 & 80.83 & 55.63 & 40.20  & 42.75     & 37.17   & 27.03   & 17.11 & 24.96   & 62.92        & 56.72          & 64.30  & 52.80 & 87.73   & 31.46  \\
    0.50          & 68.33   & 53.24 & 6.50 & 19.81 & 72.15 & 23.21 & 23.55  & 25.81     & 18.32   & 13.54   & 6.13 & 18.00   & 35.42        & 18.02          & 51.02  & 45.72 & 76.93   & 8.98  \\
    \bottomrule
    \end{tabular}
    }
\end{table*}

\myparagraph{Quantitative Evaluation on 3D Dense Captioning with LLM-Generated Boxes.} 
We evaluate the quality of the 3D bounding boxes generated by \modelname{} on 3D Dense Captioning.
Following existing works \cite{chen2021scan2cap}, we evaluate with the $m@k$IoU metric, where $m$ could be one of CiDEr, BLEU-4, METEOR, Rouge-L:
\begin{equation}
    m@k\text{IoU}\left(b^{3D}_{\text{pred}}, c_{\text{pred}}\right) 
    = \frac{1}{N}\sum_{i=1}^{N} 
        m\left(c_{\text{pred}}, c_{\text{gt}}\right) 
        \cdot
        \mathbb{I}\left\{\text{IoU}\left(b^{3D}_{\text{pred}}, b^{3D}_{\text{gt}}\right) \ge k\right\}.
\end{equation}
Here, $N$ is the total size of annotated instances, $c_{\text{pred}}$ and $c_{\text{gt}}$ are the predicted caption and the ground truth human annotations for this instance, $b^{3D}_{\text{pred}}$ and $b^{3D}_{\text{gt}}$ are the predicted 3D bounding box and the ground truth 3D box annotation,
and $\mathbb{I}\left\{ \text{condition} \right\}$ is the identification function that equals $1$ if the condition meets, and $0$ otherwise.

\begin{figure*}[htbp]
	\centering
	\includegraphics[width=0.98\linewidth]{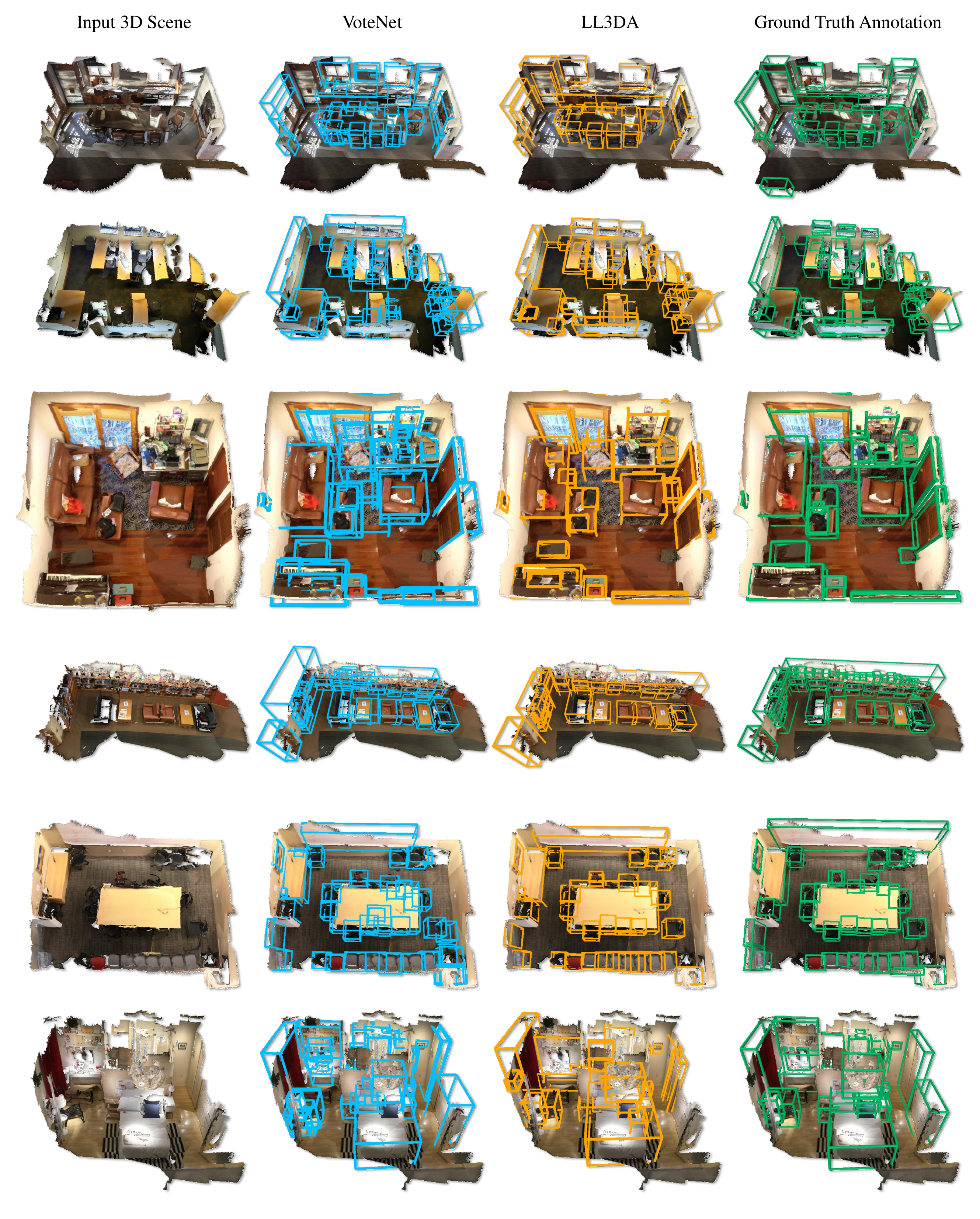}
    \vspace{-5mm}
	\caption{
    	\textbf{Visualization of the Bounding Boxes.}
        We simulate the user click with the spatial position of vote queries proposed in Vote2Cap-DETR \cite{chen2023vote2cap-detr}.
        The bounding boxes from LL3DA are reconstructed via decoding the ``\obj{$c_x$, $c_y$, $c_z$, $w$, $h$, $l$}'' from the generated texts.
        The reconstructed bounding boxes are tight to the objects in the 3D scene.
    }
	\label{appendix-fig:llm-generated-boxes}
\end{figure*}

Our baseline method is listed in Row 1 (\cref{appendix-tab:scanrefer llm boxes}).
The baseline method is also the \modelname{} model, but it only generates instance captions in response to the box prompts generated by the 3D detector \cite{chen2023vote2cap-detr} and the textual instruction ``\underline{\textit{\textcolor[RGB]{215,48,39}{\#\#\# human:} describe this object in the given 3D scene. \textcolor[RGB]{69,117,180}{\#\#\# assistant:}}}''.
The results listed in the second row of \cref{appendix-tab:scanrefer llm boxes} come from reconstructing 3D bounding boxes from decoding the ``\obj{$c_x$, $c_y$, $c_z$, $w$, $h$, $l$}'' in the generated texts.
We use ``\underline{\textit{\textcolor[RGB]{215,48,39}{\#\#\# human:} given the 3D scene, localize and describe this object. \textcolor[RGB]{69,117,180}{\#\#\# assistant:}}}'' as the textual instruction.
Though there is still gap between the bounding boxes generated by \modelname{} and 3D localizer specialists, \modelname{} could generate reasonable bounding box estimations.
One can also refer to \cref{appendix-fig:llm-generated-boxes} for more visualization details.

\begin{table*}[htbp]
    \caption{
        \textbf{3D Dense Captioning Performance with 3D Bounding Boxes Generated by \modelname{}.}
        Though there is still gap between \modelname{} and 3D specialists for object localization, \modelname{} could generate reasonable bounding box estimations.
    }
    \label{appendix-tab:scanrefer llm boxes}
    \centering
    \resizebox{0.8\linewidth}{!}{
    \begin{tabular}{cccccccccc}
    \toprule
    \multirow{2}{*}{\begin{tabular}[c]{@{}c@{}}Test Time\\Localization\end{tabular}} & \multicolumn{4}{c}{IoU = 0.25}                                              &  & \multicolumn{4}{c}{IoU = 0.5}                                                                   \\ \cline{2-5} \cline{7-10} 
                                               & C@0.25$\uparrow$ & B-4@0.25$\uparrow$ & M@0.25$\uparrow$ & R@0.25$\uparrow$ &  & C@0.5$\uparrow$ & B-4@0.5$\uparrow$ & M@0.5$\uparrow$ & R@0.5$\uparrow$ \\ \hline
    3D Detector \cite{chen2023vote2cap-detr}   & 74.17            & 41.41              & 27.76            & 59.53            &  & 65.19            & 36.79              & 25.97           & 55.06            \\
    \modelname{}                               & 62.90            & 35.14              & 26.70            & 55.62            &  & 51.12            & 29.27              & 24.18           & 49.51            \\ \bottomrule
    \end{tabular}
}
\end{table*}

\myparagraph{Qualitative Results of LL3DA's Box Predictions.}
We adopt the spatial location of vote queries to simulate the user click (256 queries per scene), and ask the model to generate 3D bounding boxes in forms of ``\obj{$c_x$, $c_y$, $c_z$, $w$, $h$, $l$}'' as introduced in the main paper.
Then, we reconstruct the 3D bounding boxes, and visualize them in in \cref{appendix-fig:llm-generated-boxes}.
The visualization results show that \modelname{} could produce tight bounding boxes close to the objects in diverse and complex 3D environments.

\clearpage

\section{Data and Instructions}
\label{appendix-sec:data}

In this section, we provide statistic details on the 3D vision and language datasets we train the model (\cref{subsec:statistics}) as well as the several textual instruction samples for different tasks (\cref{subsec:instructions}).

\subsection{Data Statistics}
\label{subsec:statistics}

Our method is trained on the following datasets:

\myparagraph{ScanNet \cite{dai2017scannet}} is a 3D indoor dataset covering diverse 3D environments, including apartments, living rooms, kitchens, bedrooms, and so on.
ScanNet splits the dataset into 1201, 312, and 100 scenes for training, validation, and testing, repectively.

\myparagraph{ScanRefer \cite{chen2020scanrefer}} is a 3D vision-language dataset, which contains 36,665 free-form natural language descriptions on 7,875 objects from 562 scenes for training, and 9,508 descriptions on 2,068 objects from 141 scenes from ScanNet \cite{dai2017scannet} for evaluation.

\myparagraph{Nr3D} \cite{achlioptas2020referit3d} is a 3D vision-language dataset with 32,919 free-form natural language descriptions on 4,664 objects from 511 scenes for training, and 8,584 descriptions on 1,214 objects from 130 scenes from ScanNet \cite{dai2017scannet} for evaluation.

\myparagraph{ScanQA \cite{azuma2022scanqa}} is a 3D vision-language dataset.
The training set of ScanQA consists of 25,563 question-answer pairs on 562 scenes from ScanNet training set.
ScanQA further splits 141 unique 3D scenes from ScanNet validation set into two sets:

\begin{itemize} 
\setlength\itemsep{0em}
    \item The validation set contains 4,675 question-answer pairs on 71 out of 141 validation scenes.
    \item The ``test w/ object'' set (``test set with ScanNet object annotations'') contains 4,976 questions on the rest 70 scenes.
\end{itemize}
Additionally, ScanQA annotates 6,149 questions on 97 scenes on the ScanNet test set as the ``test w/o object'' set (``test set without ScanNet object annotations'').

\myparagraph{3D-LLM \cite{hong20233d-llm}.} The ScanNet subset of 3D-LLM covers 1) 1,033 descriptions on 517 scenes, 2) 1,864 lines of embodied task planning on 510 scenes, and 3) 2,955 lines of multi-turn embodied dialogues on 517 scenes.
%
All of the annotated scenes come from the training set of ScanNet.
The evaluation details could be found in \cref{appendix-sec:more-evaluations}.

\subsection{Training Samples}
\label{subsec:instructions}

We list a couple of the training samples for different tasks in \cref{appendix-tab:instructions}.
The special tokens ``[Caption]'', ``[Question]'', ``[Answer]'', ``[Box]'', and ``[Response]'' will be replaced with the language annotations and 3D coordinates.

\begin{table*}[htbp]
    \caption{
        \textbf{Data Samples Used to Train \modelname{}.}
        We list a couple of data samples used to train \modelname{} for each task.
    }
    \label{appendix-tab:instructions}
    \centering
    \resizebox{\linewidth}{!}{
    \begin{tabular}{lp{10cm}lp{5cm}}
    \toprule
    Task Name & Text Instructions & Visual Interactions & Expected Output \\ \hline
    \multirow{3}{*}{3D Dense Captioning}    
            & ``\textcolor[RGB]{215,48,39}{\#\#\# human:} describe this object in the given 3D scene. \textcolor[RGB]{69,117,180}{\#\#\# assistant:} & [Click] or [Box]                & [Caption]       \\
            & ``\textcolor[RGB]{215,48,39}{\#\#\# human:} given the 3D scene, localize and describe this object. \textcolor[RGB]{69,117,180}{\#\#\# assistant:}''   & [Click] or [Box]         & the object is localized at [Box], [Caption] \\ 
    \hline
    \multirow{4}{*}{3D Question Answering}  
            & ``\textcolor[RGB]{215,48,39}{\#\#\# human:} given the 3D scene, answer the question: ``[Question]'' \textcolor[RGB]{69,117,180}{\#\#\# assistant:}''              & Optional [Clicks]   & [Answer].     \\
            & ``\textcolor[RGB]{215,48,39}{\#\#\# human:} answer the question: "[Question]" with the related object locations in the input 3D scene. \textcolor[RGB]{69,117,180}{\#\#\# assistant:}''                  & Optional [Clicks]   & the related objects are localized at [Box]. the answer is: [Answer]. \\
    \hline
    Scene Description      
            & ``\textcolor[RGB]{215,48,39}{\#\#\# human:} describe this 3D scene \textcolor[RGB]{69,117,180}{\#\#\# assistant:}''            & -  & [Caption]       \\ 
    \hline
    \multirow{5}{*}{Embodied Planning}
            & ``\textcolor[RGB]{215,48,39}{\#\#\# human:} I want to prepare and cook a meal in the kitchen. I have done these things: 1. go to the refrigerator. What should I do next? \textcolor[RGB]{69,117,180}{\#\#\# assistant:}''               & -                & open the refrigerator  \\ 
            & ``\textcolor[RGB]{215,48,39}{\#\#\# human:} I want to set up a home office workspace. What should I do? \textcolor[RGB]{69,117,180}{\#\#\# assistant:}''  & -                   & \begin{tabular}[c]{@{}l@{}}1. place a desk against the wall\\ 2. position a chair at the desk...\end{tabular}          \\ 
    \hline
    \multirow{2}{*}{Embodied Dialogue}
            & ``\textcolor[RGB]{215,48,39}{\#\#\# human:} ... \textcolor[RGB]{69,117,180}{\#\#\# assistant:}''                  & -               & [Response]        \\
            & ``\textcolor[RGB]{215,48,39}{\#\#\# human:} ... \textcolor[RGB]{69,117,180}{\#\#\# assistant:} ... \textcolor[RGB]{215,48,39}{\#\#\# human:} ... \textcolor[RGB]{69,117,180}{\#\#\# assistant:}''                  & -               & [Response]        \\
    \bottomrule
    \end{tabular}
}
\end{table*}

\clearpage

\section{Additional Implementation Details}
\label{appendix-sec:implementation}

\myparagraph{Scene Encoder.}
As introduced in the main paper, we adopt the pre-trained masked transformer encoder \cite{misra2021-3detr} as the scene encoder.
The masked transformer encoder first tokenizes the input point cloud $PC$ into 2,048 points tokens uniformly scattered in the 3D scene with a set-abstraction layer \cite{qi2017pointnet++}.
Following that are three cascaded transformer encoder blocks with masking radius of $0.16$, $0.64$, $1.44$, respectively.
There is another set-abstraction layer between the first two transformer blocks, which further down-samples the encoded tokens into 1,024 point tokens.
The output feature of the scene encoder is $f_{enc} = \mathbb{R}^{1,024 \times 256}$, representing a 256 dimensioned feature for each of the 1,024 point tokens.

\myparagraph{Multi-Modal Transformer.}
The feature dimension of the Multi-Model Transformer is 768.
We use up to 12 heads for each attention layer, and a total of six transformer layers in practice.
As mentioned in the main paper, we choose to load the pre-trained word embeddings and positional embeddings from BERT \cite{devlin2018bert}.
The total vocabulary size is 30,522, and the number of position embeddings is 512.

\myparagraph{Trainable Parameters.}
The total number of trainable parameters is about 111M, which is less than 10\% of the parameters in the frozen LLM backbone (OPT-1.3B).

\end{document}